\documentclass{article}
\usepackage{ijcai25}

\usepackage{times}
\usepackage{soul}
\usepackage{url}
\usepackage[hidelinks]{hyperref}
\usepackage[utf8]{inputenc}
\usepackage[small]{caption}
\usepackage{graphicx}
\usepackage{amsmath}
\usepackage{amssymb}
\usepackage{amsthm}
\usepackage{pifont}
\usepackage[table]{xcolor}
\usepackage{booktabs}
\usepackage{algorithm}
\usepackage{algorithmic}
\usepackage[switch]{lineno}
\usepackage{tikz}
\usepackage[inkscapelatex=false]{svg}
\usepackage{array}
\usepackage{makecell}
\usepackage{pgfplots}
\usepackage{appendix}

\title{Using Code Generation to Solve Open Instances of \\Combinatorial Design Problems}

\author{
    Christopher D. Rosin
    \affiliations
    https://constructive.codes
    \emails
    christopher.rosin@gmail.com
}

\begin{document}

\maketitle

\begin{abstract}
The \textit{Handbook of Combinatorial Designs} catalogs many types of combinatorial designs, together with lists of open instances for which existence has not yet been determined. 
 We develop a constructive protocol \textit{CPro1}, which uses Large Language Models (LLMs) to generate code that constructs combinatorial designs and resolves some of these open instances. The protocol starts from a definition of a particular type of design, and a \textit{verifier} that reliably confirms whether a proposed design is valid. The LLM selects strategies and implements them in code, and scaffolding provides automated hyperparameter tuning and execution feedback using the verifier. Most generated code fails, but by generating many candidates, the protocol automates exploration of a variety of standard methods (e.g. simulated annealing, genetic algorithms) and experimentation with variations (e.g. cost functions) to find successful approaches. Testing on 16 different types of designs, CPro1 constructs solutions to open instances for 6 of them: Symmetric and Skew Weighing Matrices, Equidistant Permutation Arrays, Packing Arrays, Balanced Ternary Designs, and Florentine Rectangles.

\end{abstract}

\section{Introduction}
\label{sec:introduction}
A \textit{Packing Array} is one of many types of combinatorial designs cataloged in the \textit{Handbook of Combinatorial Designs} \cite{handbook} (henceforth \textit{Handbook}).  A Packing Array is an $N$ by $k$ array of elements from $\{0,1,\ldots,v-1\}$, such that every pair of columns contains each ordered pair of elements at most once (Fig.~\ref{fig:examplepacking}).  Given $k$ and $v$, we want to know the largest $N$ for which a Packing Array exists (since we can always construct Packing Arrays with smaller $N$ by removing rows).  For $k=14$ $v=9$, the \textit{Handbook} lists $N=18$ as the best known.  Later results using SAT solvers raised this to $N=20$ \cite{packingsat}.  The upper bound is $N \leq 21$ \cite{handbook}, so it is an open question whether $N=21$ is possible.

\begin{figure}[h!]
    \centering
    \includegraphics[width=1 \linewidth]{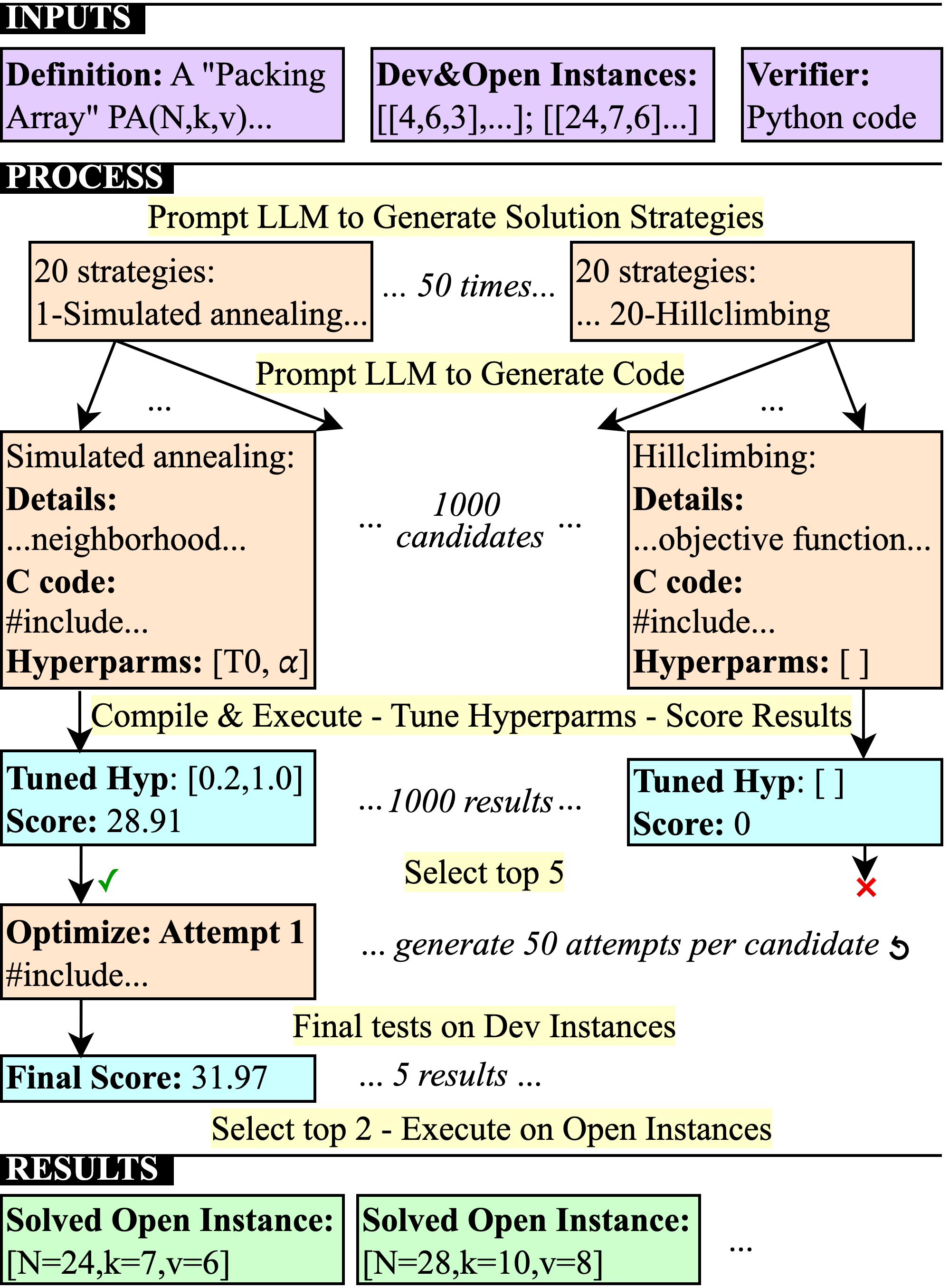}
    \caption{\textbf{Constructive Protocol CPro1}}
    \label{fig:protocol}
\end{figure}

One approach involves trying various heuristic methods, experimentally tuning each to determine whether they can construct the desired design.  In this paper, we develop a protocol \textit{CPro1} (Fig.~\ref{fig:protocol}) that uses LLMs to generate code for diverse candidate methods. The protocol automates an experimental process to identify and optimize heuristic construction strategies. It succeeds in constructing a Packing Array with $N=21$ for $k=14$ $v=9$, fully resolving this open question.  The protocol also resolves open instances of other existence problems from the \textit{Handbook}.  Tables~\ref{tab:mainresults}~and~\ref{tab:prototpyingresults} show the main results.

 \begin{table*}[t!]
     \centering
     \begin{tabular}{@{}m{8.23cm} >{\centering}m{3.2cm} | >{\centering}m{3.1cm} | m{1.4cm}@{} }
        \rowcolor{yellow!15} \centering \mbox{Combinatorial design problem \& definition} \mbox{(as used by CPro1)} & \textit{Handbook}'s smallest open instances & Progress on these, from the literature & \textbf{This \mbox{paper}} \\  [-0.25ex]
      \end{tabular}
      \begin{tabular}{@{}m{8.3cm} l@{}}
                \hline

          \textbf{Packing \mbox{Array} (PA)}:  A "Packing Array" PA(N,k,v) is an N x k array (N rows and k columns), with each entry from the v-set    
 \{0,1,...v-1\}, so that every N x 2 subarray contains every ordered pair of symbols at most once.  Given (N,k,v), we want to construct PA(N,k,v). & {\rowcolors{3}{black!8}{black!1}
\begin{tabular}{ >{\centering}m{3.2cm} | >{\centering}m{3.1cm} | m{1.4cm}}
\rowcolor{green!10}  \mbox{Bounds on max N} \mbox{for each (k,v)} & \mbox{\cite{stardom}} \mbox{\cite{packingsat}} & CPro1 \\ 
\hline

$(5,6)$: $34 \geq $ N $ \geq 30$ & N $ \geq 31$ &  \\
$(6,6)$: $34 \geq $ N $ \geq 30$ & N $ \geq 31$ & \\
$(7,6)$: $34 \geq $ N $ \geq 16$ & N $ \geq 23$ & \textbf{N} \boldmath{$\geq 24$} \\
$(8,6)$: $19 \geq $ N $ \geq 12$ & N $ \geq 17$ & \textbf{N} \boldmath{$\geq 18$} \\
$(9,6)$: $14 \geq $ N $ \geq 12$ & N $ = 14$ &  \\
$(11,7)$: $15 \geq $ N $ \geq 14$ & N $ = 15$ & \\
$(10,8)$: $34 \geq $ N $ \geq 22$ & N $ \geq 25$ & \textbf{N} \boldmath{$\geq 28$}  \\
$(11,8)$: $25 \geq $ N $ \geq 16$ & N $ \geq 22$ & \textbf{N} \boldmath{$\geq 24$}  \\
$(12,8)$: $19 \geq $ N $ \geq 16$ & N $ = 19$ &   \\
$(13,8)$: $17 \geq $ N $ \geq 16$ & N $ = 17$ &   \\
$(11,9)$: $45 \geq $ N $ \geq 27$ & N $ \geq 29$ &  \textbf{N} \boldmath{$\geq 32$}  \\
$(12,9)$: $30 \geq $ N $ \geq 27$ & & \textbf{N} \boldmath{$ \geq 28$}  \\
$(14,9)$: $21 \geq $ N $ \geq 18$ & N $ \geq 20$ & \textbf{N} \boldmath{$ = 21$}  \\
$(15,9)$: $19 \geq $ N $ \geq 18$ &  & \textbf{N} \boldmath{$ = 19$}  \\

\end{tabular}
} \\
          \hline

         \textbf{Skew \mbox{Weighing} \mbox{Matrix} (SkewW)}: A "weighing matrix" W(n,w) with parameters (n,w) is an n by n square matrix (n rows and n columns) with entries in \{0,1,-1\} that satisfies W W\textasciicircum T = wI.  That is, W times its transpose is equal to the constant w times the identity matrix I.  The weighing matrix will have w nonzero entries in each row and each column.  And each pair of distinct rows is orthogonal (dot product zero).  Given (n,w), we want to construct "SkewW", a skew weighing matrix W(n,w) that satisfies these properties and is also a skew matrix: that is, W\textasciicircum T = -W. & {\rowcolors{3}{black!8}{black!1}
\begin{tabular}{>{\centering}m{3.2cm} | >{\centering}m{3.1cm} | m{1.4cm}}
\rowcolor{green!10} \mbox{\phantom{xxx}Open instances\phantom{xxx}} \mbox{(n,w)}& -- & CPro1 \\ 
\hline
(18,9)? & & \textbf{Exists} \\
(30,25)? & & \\
 \makecell{ \\ \\ \\ \\ \\ \\} & & \\
\end{tabular}
} \\
          \hline

          \textbf{Balanced Ternary \mbox{Design} (BTD)}: A "Balanced Ternary Design" BTD(V,B;p1,p2,R;K,L) is an arrangement of V elements into B multisets, or blocks, each of cardinality K (K\textless=V) satisfying: 1. Each element appears R=p1 + 2*p2 times altogether, with multiplicity one in exactly p1 blocks and multiplicity two in exactly p2 blocks. 2. Every pair of distinct elements appears L times; that is, if m\_\{vb\} is the multiplicity of the v'th element in the b'th block, then for every pair of distinct elements v and w, sum\_\{b=1\}\textasciicircum      \{B\} m\_\{vb\} m\_\{wb\} = L.  The BTD is represented by a V by B incidence matrix with elements in           \{0,1,2\}.  The matrix element m\_\{vb\} in the v'th row and b'th column is the multiplicity of the v'th element in the b'th block.  The sum of each row is R, and the sum of each column is K. Given (V,B,p1,p2,R,K,L) we want to find BTD(V,B;p1,p2,R;K,L). &  {\rowcolors{3}{black!8}{black!1}
\begin{tabular}{>{\centering}m{3.2cm} | >{\centering}m{3.1cm} | m{1.4cm}}
\rowcolor{green!10}  \mbox{Open instances} \mbox{(V,B;p1,p2,R;K,L)}& \cite{designs2002} & CPro1                              \\ 
\hline
(14,18;7,1,9;7,4)? &  & \\
(12,15;6,2,10;8,6)? &  & \\
(12,20;4,3,10;6,4)? &  & \\
(16,22;9,1,11;8,5)? &  &\\
(17,17;8,2,12;12,8)? & & \textbf{Exists} \\
(14,21;6,3,12;8,6)? &  & \textbf{Exists} \\
(12,16;4,4,12;9,8)? &  & \textbf{Exists} \\
(12,26;3,5,13;6,5)? & Doesn't Exist & \\
(16,16;7,3,13;13,10)? & & \textbf{Exists} \\
(12,21;4,5,14;8,8)? & & \textbf{Exists} \\
(12,28;10,2,14;6,6)? & Exists  & \\
(14,28;8,3,14;7,6)? &   &\\
(18,18;2,6,14;14,10)? & Exists  &  \\
\end{tabular}
}  \\
          \hline
                   \textbf{Florentine \mbox{Rectangle} (FR)}:A "Florentine Rectangle" FR(r,n) is an r x n array (r rows and n columns), with each row having a permutation of the set of symbols S=\{0,1,2,...,n-1\}, such that for any two distinct symbols a and b in S and each m in       
 \{1,2,3,...,n-1\} there is at most one row in which b appears in the position which is m steps to the right of a.  A single row will have n-m pairs of symbols a,b with b being m steps to the right of a; so n-1 pairs with b directly to the right of a, n-2 with b 2 steps to the right of a, and only 1 pair with b n-1 steps to the right of a.  Given (r,n) we want to construct a FR(r,n). &   {\rowcolors{3}{black!8}{black!1}
\begin{tabular}{>{\centering}m{3.2cm} | >{\centering}m{3.1cm} | m{1.4cm}}
\rowcolor{green!10}  \makecell{Bounds on max r\\for each n} & -- & CPro1\\ 
\hline
\phantom{n}n $= 13$: $13 \geq$ r $\geq 12$ & & \\
n $= 14$: $14 \geq$ r $\geq 7$ & & \\
n $= 15$: $15 \geq$ r $\geq 7$ & & \\
n $= 20$: $20 \geq$ r $\geq 6$ & & \textbf{r} \boldmath{$\geq 7$} \\
n $= 21$: $21 \geq$ r $\geq 7$ & & \\
n $= 24$: $24 \geq$ r $\geq 6$ & & \textbf{r} \boldmath{$\geq 7$} \\
n $= 25$: $25 \geq$ r $\geq 6$ & & \textbf{r} \boldmath{$\geq 7$} \\
n $= 26$: $26 \geq$ r $\geq 6$ & & \textbf{r} \boldmath{$\geq 7$} \\
n $= 27$: $27 \geq$ r $\geq 6$ & & \textbf{r} \boldmath{$\geq 7$} \\

\end{tabular}
}\\
         \hline
         
           \end{tabular}
     \caption{\textbf{Open Instance Results}.  For each combinatorial design problem, the smallest open instances from the \textit{Handbook} are listed, along with progress on these published elsewhere in the literature.  The protocol CPro1 in Fig.~\ref{fig:protocol} used the given definition as input to generate code, and this code then made further progress on the open instances, as indicated in the last column.  The designs constructed by the code are shown in Appendix~\ref{sec:solutions}. }
     \label{tab:mainresults}
 \end{table*}

 \begin{table*}[t!]
     \centering
     \begin{tabular}{@{}m{8.23cm} >{\centering}m{3.2cm} | >{\centering}m{3.1cm} | m{1.4cm}@{} }
        \rowcolor{yellow!15} \centering \mbox{Combinatorial design problem \& definition} \mbox{(as used by CPro1)} & \textit{Handbook}'s smallest open instances & Progress on these, from the literature & \textbf{This \mbox{paper}} \\  [-0.25ex]
      \end{tabular}
      \begin{tabular}{@{}m{8.3cm} l@{}}
                \hline
         \textbf{Symmetric \mbox{Weighing} Matrix (SymmW)}: A "weighing matrix" W(n,w) with parameters (n,w) is an n by n square matrix (n rows and n columns) with entries in \{0,1,-1\} that satisfies W W\textasciicircum T = wI.  That is, W times its transpose is equal to the constant w times the identity matrix I.  The weighing matrix will have w nonzero entries in each row and each column.  And each pair of distinct rows is orthogonal (dot product zero).  Given (n,w), we want to construct "SymmW", a symmetric weighing matrix W(n,w) that satisfies these properties and is also a symmetric matrix. & 
  
         {\rowcolors{3}{black!8}{black!1}
\begin{tabular}{ >{\centering}m{3.2cm} | >{\centering}m{3.1cm} |m{1.4cm}}
          
\rowcolor{green!10}  Open (n,w)& \mbox{\cite{georgiou}} \mbox{\cite{dinitzupdate5}} & CPro1 \\ 
\hline
(14,9)? & Exists & \\
(19,9)? & & \textbf{Exists} \\
(21,9)? & & \textbf{Exists} \\
(22,16)? & & \\
(23,16)? & Exists & \\
(25,16)? & & \\
(27,16)? & & \\
(29,16)? & & \\
(28,25)? & & \\
(30,25)? & Exists &  \\
\end{tabular}
} \\
          \hline
                   \textbf{Equidistant \mbox{Permutation} \mbox{Array}~(EPA)}: An "equidistant permutation array" (EPA) with parameters (n,d,m) can be represented as an m by n matrix (m rows and n columns), where each row is the permutation of the numbers 0 to n-1.  Each pair of distinct rows must differ in exactly d positions. Given (n,d,m), we want to construct an equidistant permutation array (EPA) with these parameters. &   {\rowcolors{3}{black!8}{black!1}
\begin{tabular}{>{\centering}m{3.2cm} | >{\centering}m{3.1cm} | m{1.4cm}}
\rowcolor{green!10}  \makecell{Bound on max m\\for each (n,d)} & -- & CPro1\\ 
\hline
$(10,7)$: m $\geq 17$ & & \\
$(11,7)$: m $\geq 17$ & &\\
$(9,8)$: m $\geq 20$ & &\\
$(10,8)$: m $\geq 20$ & &\\
$(11,8)$: m $\geq 20$ & &\\
$(12,8)$: m $\geq 20$ & & \textbf{m} \boldmath{$\geq 21$} \\

\end{tabular}
}\\

  \hline
         
           \end{tabular}
     \caption{\textbf{Open Instance Results - Prototyping Set}.  This has the same format as Table~\ref{tab:mainresults}, and the results in the last column were generated by the automated protocol CPro1.  These instances were in the prototyping set, and so were initially solved in this paper via manually developed local search methods.  The designs constructed by the CPro1-generated code are shown in Appendix~\ref{sec:solutions}. }
     \label{tab:prototpyingresults}
 \end{table*}

\begin{figure}[h!]
\centering
\[
\fbox{
$\begin{array}{*{6}{r}}
    2 & 0 & 1 & 1 & 1 & 0\\
    0 & 1 & 0 & 2 & 1 & 2\\
    0 & 2 & 1 & 0 & 0 & 1\\
    1 & 0 & 2 & 0 & 2 & 2
\end{array}$}
\]
\caption{\textbf{Packing Array example} with N=4 k=6 v=3}
\label{fig:examplepacking}
\end{figure}

\begin{table*}[h]
\centering
{\rowcolors{3}{black!8}{black!1}
\begin{tabular}{l c c c}
\textbf{Combinatorial Design} & \textit{Handbook} Chapter & Solved Open Instances & Prototyping Set \\
\hline
Equidistant Permutation Array (EPA) & VI.44 & Yes & Yes \\
Symmetric Weighing Matrix (SymmW, or SyW) & V.2 & Yes & Yes \\
Skew Weighing Matrix (SkewW, or SkW) & V.2 & Yes &  \\
Packing Array (PA) & III.3 & Yes &  \\
Balanced Ternary Design (BTD) & VI.2 & Yes & \\
Florentine Rectangle (FR) & VI.62 & Yes & \\
Balanced Incomplete Block Design (BIBD) & II.1 & & \\
Bhaskar Rao Design  & V.4 & & \\
Supersimple Design & VI.57 & & \\
Difference Triangle Set &  VI.19 & &\\
Costas Array & VI.9 & &\\
Tuscan-2 Square & VI.62 & & \\
Circular Florentine Rectangle & VI.62 & &\\
Coverings & VI.11 & &\\
Perfect~\mbox{Mendelsohn}~\mbox{Design} & VI.35 & &\\
Orthogonal Array & III.6 & &  \\
\hline
\end{tabular}
}
\caption{\textbf{Selected Combinatorial Designs} Column \textit{Solved Open Instances} indicates designs for which the automated protocol CPro1 succeeded in solving at least one open instance.  Column \textit{Prototyping Set} indicates those for which the open instances were first solved via manually created local search.}
\label{tab:selected}
\end{table*}

\begin{table}[h]
 {\rowcolors{3}{black!8}{black!1}
\begin{tabular}{m{1.6cm} m{1.7cm} m{0.7cm} m{1.4cm} m{1.0cm}}
\mbox{\phantom{placehold}} \mbox{\phantom{placehold}} LLM & \mbox{\phantom{placehold}} \mbox{\phantom{placehold}} Ver. Date & \mbox{\phantom{place}} num. cand. & \centering SymmW open (of~2) & EPA open (of~1) \\
\hline
GPT-4o & 2024-05-13 & \raggedleft 100 & \centering 0 &  1 \\
GPT-4o & 2024-05-13 & \raggedleft 300 & \centering 1 & 0 \\
GPT-4o & 2024-05-13 & \raggedleft 1000 & \centering 2 & 1 \\
Claude 3.5 Sonnet & 20241022 & \raggedleft 1000 & \centering 2 &  0 \\
Mistral Large & 2407 & \raggedleft 1000 & \centering 2 & 0 \\
DeepSeek v2.5 & 2024/09/05 & \raggedleft 1000 & \centering 0 & 0 \\
Qwen2.5-\mbox{72B-Instruct} & 11-2024 &  \raggedleft 1000 &  \centering 0 & 0 \\
\hline
\end{tabular}
}
\caption{\textbf{Prototyping Results}: For each Large Language Model and total number of candidate programs, the table shows the number of distinct open instances from the prototyping set that were solved by the run.}
\label{tab:models}
\end{table}

\begin{table}[h]
\begin{tabular}{|m{2.5cm}|m{0.5cm}|m{0.5cm}|m{0.5cm}|m{0.5cm}|m{0.5cm}|m{0.5cm}|}
\hline
 & PA & SyW & SkW & BTD & FR & EPA \\
\hline
\textbf{GPT-4o} & \cellcolor{lime}cSA &\cellcolor{lime} rSA & \cellcolor{lime} cSA & \cellcolor{lime} GA& \cellcolor{lime} DFS & \cellcolor{lime}cSA \\
\hline
- Reduce runtime  & \cellcolor{lime}cSA & \cellcolor{lime} rSA & \cellcolor{lime} cSA  & \cellcolor{lime} GA & \cellcolor{lime} DFS  & \\
\hline
- No final dev test & \cellcolor{lime}cSA  & \cellcolor{lime} rSA  &  \cellcolor{lime} cSA  & \cellcolor{lime} GA  & \cellcolor{lime} DFS  & \\
\hline
- No optimization & \cellcolor{lime}cSA & \cellcolor{lime} SA & \cellcolor{lime} cSA & \cellcolor{lime} GA  & & \\
\hline
- No hyper tuning   & \cellcolor{lime}rSA & \cellcolor{lime}GA & & & & \\
\hline
\hline
\textbf{Mistral Large}  & \cellcolor{lime}cSA & \cellcolor{lime} cSA  & \cellcolor{lime} cSA  & & & \\
\hline
- Reduce runtime  & \cellcolor{lime}cSA  & \cellcolor{lime} cSA & \cellcolor{lime} cSA & & & \\
\hline
- No final dev test & \cellcolor{lime}cSA & \cellcolor{lime} cSA &  \cellcolor{lime} cSA  & & & \\
\hline
- No optimization & \cellcolor{lime}cSA  &  & \cellcolor{lime}cSA & & & \\
\hline
- No hyper tuning  & \cellcolor{lime}cSA &  & & & & \\
\hline
\end{tabular}
\caption{\textbf{Experiment Results} Each row is an experiment, each column a type of combinatorial design.  Green cells show where the experiment solved at least one open instance of that combinatorial design problem.  The first set of rows use GPT-4o, and the ``-" rows stack up successive ablations: \textbf{Reduce runtime} reduces from 48 hours to 2 for open instances, \textbf{No final dev test} eliminates final 2 hour testing on development instances (instead using original 50-second test results), \textbf{No optimization} eliminates the code optimization step, and \textbf{No hyper tuning} eliminates hyperparameter tuning (instead using default hyperparameters given by the LLM).  The second set of rows use Mistral Large.  Each cell shows the strategy that obtained success: \textbf{DFS}: backtracking depth-first search, \textbf{GA}: genetic algorithm, \textbf{SA}: simulated annealing with a slow cooling schedule, \textbf{rSA}: simulated annealing with periodic resets, \textbf{cSA}: constant-temperature simulated annealing.}
\label{tab:successes}
\end{table}

\section{Related Work}

Code generation is one of the primary applications of LLMs \cite{codegen,codegen2}.  LLMs have the potential to translate natural language requirements to working programs, even without being given the ability to fully test the code.  

For difficult problems, it can help generate multiple candidate programs and test whether any succeed \cite{passatk}. Protocols that prompt the LLM for a natural-language description before generating code can generate more diverse and successful candidates \cite{plansearch}, and we use such an approach here.  

In code generation benchmarks, LLM-generated code is commonly tested on a small number of test cases.  Particularly when generating multiple candidate solutions, this can lead to a situation where generated code is correct on the limited set of test cases, but fails to generalize to other inputs \cite{passatk2}.  Only a restricted set of problems, for which an \textit{oracle verifier} can fully verify solutions, benefit from generating arbitrarily large numbers of candidates \cite{oracleverifier}.  In this paper, our problems have such oracle verifiers, so we can fully benefit from generating large numbers of candidate solutions.

LLMs have been applied to rewrite existing code to improve its performance \cite{llmcodeopt}, including by sampling and testing multiple optimization candidates from the LLM \cite{llmcodeoptcandidates}.  Our protocol includes a step that attempts to improve performance by generating multiple optimization candidates and measuring their performance.

Code generation with LLMs has been used to develop and improve heuristics for combinatorial optimization problems \cite{eoh}, including generating improved search operators for genetic algorithms \cite{reevo}.  In this paper, we use LLMs to propose and implement heuristic strategies in an open-ended way, and the successful strategies that emerge include genetic algorithms \cite{ga} and simulated annealing \cite{sa}.

LLaMEA-HOP \cite{automatedhyperparameter} uses automated hyperparameter tuning together with LLM-generated heuristics.  The authors motivate it by noting it is potentially very costly to use the LLM for hyperparameter tuning, so they ask the LLM to expose a hyperparameter configuration space, and then offload hyperparameter tuning to a separate specialized system.  We share this motivation, and use automated hyperparameter tuning for generated code that exposes hyperparameters. 

Efforts to apply LLMs to mathematics have focused on benchmarks with known solutions \cite{math,olympiad} and generation of step-by-step proofs that could be verified with systems like Lean \cite{leanllm,oracleverifier}.  Here, we focus on problems that can be resolved by constructing a combinatorial object that can be easily verified, rather than requiring a step-by-step proof.  This has the potential to resolve open questions in mathematics without the difficulty of generating lengthy step-by-step proofs.

FunSearch \cite{funsearch} used LLMs as part of an evolutionary algorithm that searched for greedy functions to construct Cap Sets, and succeeded in constructing a Cap Set of size 512 for dimension $n=8$, a result which addresses a recognized open question.  Like combinatorial designs, Cap Sets are combinatorial objects that can be readily tested by an oracle verifier.  Compared to FunSearch, our work here generates far fewer candidate programs (thousands rather than millions), and allows open-ended strategies which can use much more compute during execution than greedy functions.

\section{Method}

\subsection{Terminology}
\textit{Combinatorial designs} are systems of finite sets that satisfy a set of constraints.  The specific finite sets and constraints involved define the {\em type} of combinatorial design (e.g. Packing Arrays, or Balanced Incomplete Block Designs).  The \textit{existence problem} (or \textit{combinatorial design problem}) for a particular type of combinatorial design has a small number of input \textit{parameters} (e.g. size), and asks whether it is possible to construct a design that satisfies the constraints for these parameters.  An \textit{instance} of the combinatorial design problem specifies particular numerical values for the parameters, and a \textit{solution} to the instance would exhibit a combinatorial object (the \textit{design}) that meets the constraints instantiated with these parameters (or a proof that none exists, but in this paper we are limited to seeking solutions that construct the design).  For example, one instance of the existence problem for Packing Arrays would ask whether a design with $N=21$ $k=14$ $v=9$ exists, and a solution could give an $N$ by $k$ array that meets the constraints of the definition given in Section~\ref{sec:introduction}.  Note there may be more than one solution to a particular instance of the existence problem.  

A \textit{Large Language Model} (LLM) takes a textual \textit{prompt} as input and returns a textual response, which may include natural language and/or programming language code.  LLMs are trained via machine learning, but in this paper we are only using off-the-shelf pretrained LLMs.  We use the term \textit{protocol} to refer to an algorithm or system which includes calls to an LLM, and the \textit{scaffolding} consists of the elements of this protocol other than the LLM.

\subsection{Selection of Combinatorial Designs}
From the \textit{Handbook}, we select 16 types of combinatorial designs (Table~\ref{tab:selected}) that have clearly defined open instances with relatively small parameters that might be amenable to heuristic search.  For each of these, we review the literature to identify which instances have remained open.

For each selected type of design, we provide:
\begin{description}
\item[Textual Definition] Defines the design and mandates a specific representation as an array of small integers.
\item[Verifier in Python] Determines whether a proposed design in this representation is correct.
\item[Open Instances] Parameters for which existence of this type of design is not yet known.  The ultimate goal is to construct designs with these parameters.
\item[Development Instances (Dev Instances)] Parameters for which designs of this type are known to exist, including some of the smallest ones, as well as ones just slightly smaller than the open instances.  Candidate approaches have their generated code executed on these, with results checked by the Verifier, to identify the most promising candidates.
\end{description}

\subsection{Prototyping Set}
Our goal is to develop an LLM-based protocol which could resolve open instances of combinatorial design existence problems.  Since the existence of open instances is unknown, it could be that they don't exist and constructing them is impossible -- this makes it difficult to prototype protocols for constructing these designs.  To aid in prototyping, we first establish a small {\em prototyping set} of open instances that actually do exist and can be constructed.  We manually experimented with local search methods for 5 of of the selected combinatorial designs: Bhaskar Rao Designs, Difference Triangle Sets, Equidistant Permutation Arrays (EPA), Supersimple Designs, and Symmetric Weighing Matrices (SymmW).  The local search methods select changes which minimize a cost function, while sometimes accepting worsening moves to escape local optima.  Local search succeeded in resolving one open instance for EPA and two for SymmW; these form our prototyping set for use while developing a protocol that can also resolve these.   The manually-developed local search algorithm for EPA is shown in Algorithm~\ref{alg:epals}.

\subsection{Protocol Development}
\label{sec:methoddev}
In our protocol, an LLM takes a combinatorial design’s definition as input, selects diverse strategies, and generates code for them.  We provide scaffolding that compiles and executes the generated code on development instances, checks results with the verifier, and selects the most promising candidates for use on open instances.  We generate code in C to take advantage of its speed.

Experiments on the prototyping set used GPT-4o \cite{gpt4o} and Claude 3.5 Sonnet \cite{sonnet}.  For GPT-4o, the 2024-05-13 version gave better initial results than a later version, so we continued using the 2024-05-13 version.  Through experimentation on the prototyping set, we establish these observations and elements of the protocol:

\begin{description}
\item[Prompting for Diversity] Prompting for 20 strategies yields more diverse approaches than asking for 10, or repeatedly asking for 1.
\item[Prompting for Details, then Code] Prompting for a detailed textual description of the approach, before prompting for the code, improves reliability of the generated code.
\item[Prompting to Prevent Early Termination] We expect solutions to run for minutes or hours before succeeding. Generated code often terminates too quickly (milliseconds), which we partially mitigate by prompting for code that should “not terminate until a valid solution is found.”
\item[Multiple Random Seeds] Testing development instances on multiple seeds encourages randomized strategies, which have an increased chance of success when running multiple seeds in parallel on open instances.  We run a candidate program on all of the development instances, each with multiple seeds, in parallel (the rest of the protocol is serial).
\item[LLMs Fail to Generate Viable Local Search] Our manually developed solutions to the prototyping instances relied on local search.  While the LLMs often propose strategies that the LLM describes as ``Local Search", their actual implementations are consistently naive: simple hill climbing which quickly gets stuck in local optima, or pure random search that makes no progress.  This is concerning, particularly since local search methods which avoid local optima are well-established \cite{hoosbook,numvc}.
\item[Automated Hyperparameter Tuning] The LLMs sometimes propose simulated annealing, which could potentially serve as a viable alternative to local search.  But the LLMs usually propose naive annealing schedules that very quickly ramp temperature down to the point where it becomes simple hill climbing stuck in local optima.  We partially mitigate this by asking the LLM to expose hyperparameters and provide ranges and defaults for them, and we then have the scaffolding provide automated hyperparameter tuning.  We find extreme ends of the range are often important (e.g. simulated annealing cooling rate of 1.0, which yields constant-temperature simulated annealing that resembles local search), and we find that an adaptive hyperparameter tuning method \cite{optuna_2019} struggles to explore these extremes.  We therefore provide grid based hyperparameter tuning, that uses a coarse-grained linear grid across the middle of the range of each hyperparameter, and logarithmic/fine-grained near the end points.  Hypertuning begins with up to 1000 settings to the hyperparameters, and tests each of these settings for 0.5 seconds on the development instances and seeds in parallel (for a total of 500 seconds elapsed wall clock time, excluding overhead).  Then the next round takes up to 100 of the best settings and executes for 5 seconds, and a final round takes up to 10 of the best settings and executes for 50 seconds, returning the best-scoring setting to the hyperparameters.  This is able to find successful hyperparameter settings for simulated annealing.  
\item[Prompting for Code Optimization] Prompting for iterative refinement to generated code has limited success.  Open-ended refinements are likely to abandon what was already successful.  But more limited refinements that prompt only for optimizing speed are more successful - e.g. converting an expensive cost function implementation into a faster incremental one.  Our protocol samples 50 optimization candidates from the LLM, and then if there is substantial improvement it repeats the process (up to 5 rounds).
\item[Sandboxing] Running the generated code in a protected sandbox is essential.  Otherwise, buggy C code can allocate excessive memory that results in crashing the whole protocol.  We use \textit{firejail} on Linux, which also blocks file system and network access, and can block problematic programs without crashing the rest of the protocol.
\item[Generate Many Candidates] Most generated code fails, often for simplistic reasons (see Section~\ref{sec:failuremodes}).  But repeating the process eventually obtains code that succeeds on some development instances.  Running the protocol at smaller scale with a total of 100 or 300 generated programs failed to solve the full set of 3 prototyping instances (see Table~\ref{tab:models}), but 1000 candidates (50 repetitions of 20 strategies each) succeeded.
\end{description}

Figure~\ref{fig:protocol} outlines the final constructive protocol CPro1, with further details in Algorithm~\ref{alg:cpro1}.  During initial testing, each of the 1000 candidate programs is scored based on executing it for 50 seconds using the tuned hyperparameters.  The 5 top-scoring candidates are then selected for optimization, which also gives each candidate 50 seconds of execution.  For final testing on the development instances, the same 5 top candidates (after optimization) are given 2 hours to execute.  The 2 top-scoring candidates from this are then run for 48 hours on the open instances.

This protocol, with the choices made here, successfully solves the development instances using GPT-4o.  For EPA, it solves using a constant-temperature simulated annealing that is somewhat similar to our manually implemented local search (see Algorithm~\ref{alg:epacpro1}).  For SymmW, CPro1's solution solves the instances with simulated annealing that periodically resets the temperature to avoid getting stuck.  Claude 3.5 Sonnet succeeds on SymmW but not EPA.

We also test several open-weights LLMs on the development set (Table~\ref{tab:models}).  Of these, Mistral Large is most successful, so we also use it to provide additional results with improved reproducibility using an open-weights model.

\subsection{Experiment}

For each of the 16 types of combinatorial design, we run the protocol once with GPT-4o and once with Mistral Large.  We also test ablated versions of the protocol on each type of combinatorial design.  Each ablation uses the same candidate programs generated by the original run.  

When we succeed in solving least one open instance of a combinatorial design problem, we extend by testing the generated code on adjacent open instances, including all of the smallest open instances from the \textit{Handbook} shown in Tables~\ref{tab:mainresults}~and~\ref{tab:prototpyingresults}.

The experiments are run on a Linux machine with AMD Ryzen 9 7950X3D CPU and 128GB of memory, and the machine is fully dedicated to one experiment at a time.  
With the chosen parameters shown in Algorithm~\ref{alg:cpro1}, a full run on one type of combinatorial design takes approximately 7 days of runtime.  The majority of the runtime is used running candidate programs on development instances and open instances.

\section{Results}

Tables~\ref{tab:mainresults}~and~\ref{tab:prototpyingresults} show the main results.  The protocol CPro1 succeeds on open instances of the existence problem for 6 types of combinatorial designs, including 4 that were not in the prototyping set.  The tables show that there has been some progress elsewhere in the literature on these open instances from the \textit{Handbook}, and the results from CPro1 add significantly to this progress.

For the other types of combinatorial designs in Table~\ref{tab:selected}, code generated by CPro1 was able to solve instances in the development set, but this code was unable to solve any of the open instances.  Neither GPT-4o nor Mistral Large solved any open instances for these, and none of the ablation runs solved any open instances for these.

The successful results are constructed by programs that range from 120-270 lines of C code.

Results for both GPT-4o and Mistral Large are shown in Table~\ref{tab:successes}.  CPro1 using Mistral Large solves a subset of the open instances that are solved using GPT-4o.  The full results in Tables~\ref{tab:mainresults}~and~\ref{tab:prototpyingresults} and Appendix~\ref{sec:solutions} were generated using GPT-4o.

\subsection{Ablation}
Table~\ref{tab:successes} also shows ablation results.  All of the considered elements of the protocol, including automated hyperparameter tuning and optimization of the initial code, are needed to obtain the full results.  Automated hyperparameter tuning was developed for simulated annealing during prototyping (Sec.~\ref{sec:methoddev}), but it was also needed to obtain positive results from the genetic algorithm code generated for Balanced Ternary Designs.

\begin{algorithm}[t!]
    \caption{Manually created local search for EPA.  It can escape local optima by randomly selecting a row $r$ in which all column swaps are worsening moves; it is forced to choose one of these. \phantom{descendery}}
    \label{alg:epals}
    \textbf{Input}: N,k,d\\
    \textbf{Output}: Valid EPA with parameters (N,k,d)\\
    \\
    \textbf{State}: $E$ has $N$ rows, each a permutation of $\{0,1,\ldots,k-1\}$\\
    \\
    \phantom{de}\mbox{\textbf{def} $\delta(i,j)$: return \#columns differing in rows $i$,$j$ of $E$}\\
    \phantom{de}\textbf{def} cost = $\sum_{\mathrm{rows}\: i<j} |\delta(i,j)-d|$\\
  
    \begin{algorithmic} 
        \STATE \textbf{Initialize} $E$ with random permutation in each row
        \WHILE{cost $>0$}
        \STATE $r$ = random choice from \{rows $i: \exists j$ with $\delta(i,j)>0$\}
        \STATE $c,d$ = distinct col. in $r$ that, if swapped, minimize cost
        \STATE \textbf{swap} columns $c$ and $d$ in $r$
        \ENDWHILE
        \STATE \textbf{return} $E$
    \end{algorithmic}
\end{algorithm}

\subsection{Strategies Implemented by Generated Code}
\label{sec:strategies}

Each run has a total of 1000 candidates, starting from 50 lists of 20 proposed strategies each.  With both GPT-4o and Mistral Large, simulated annealing and genetic algorithms are frequently proposed -- often close to 50 times (nearly every list of proposed strategies) or even more than 50 if alternate names are counted.  Other frequently proposed strategies include depth-first search (under various names -- e.g. ``backtracking"), greedy algorithms, tabu search, recursive constructions, branch and bound, and more.  

All of the successes in Table~\ref{tab:successes} were due to simulated annealing or genetic algorithms, except for Florentine Rectangles which used depth-first search.  

As discussed in Sec.~\ref{sec:methoddev}, simulated annealing runs the risk of rapid cooling schedules getting permanently stuck in local optima, which destroys the productivity of the extended runs that are needed for many of these results.  The code generated here mitigated this by using a constant temperature (which is fine-tuned by automated hyperparameter tuning), or a periodic reset, or (in one ablation run) an extremely slow cooling schedule.  The constant temperature implementations were achieved by a mix of hyperparameter tuning setting cooling rate to 1.0, and generated code which hardwired a constant temperature -- however in both cases the LLM-generated strategy details do not explicitly address the value of a constant temperature; even when hardwired it seems more like an accidental feature that ends up performing well.  

Algorithms~\ref{alg:epals}~and~\ref{alg:epacpro1} compare manually developed local search (from the effort to build the prototyping set) with CPro1's automatically generated constant-temperature simulated annealing for EPA.  The approaches are similar, but have different methods of escaping local optima.  CPro1's solution performs somewhat better: on the hardware used for these experiments, for open instance n=12 d=8 m=21, CPro1's simulated annealing has a median solution time of 10 hours, compared to 21 hours for manually developed local search.

The genetic algorithm for Balanced Ternary Designs uses a rather small population of only 100, running for over a billion generations to construct some of the designs.  This genetic algorithm has no elitism, and a high mutation rate, which could maintain diversity during such extended runs.

\begin{algorithm}[t!]
    \caption{CPro1-generated simulated annealing for EPA.  It can escape local optima by accepting a worsening move, with a probability controlled by a constant temperature selected by automated hyperparameter tuning.}
    \label{alg:epacpro1}
    \textbf{Input}: N,k,d\\
    \textbf{Output}: Valid EPA with parameters (N,k,d)\\
    \\
    \textbf{State}: $E$ has $N$ rows, each a permutation of $\{0,1,\ldots,k-1\}$\\
    \\
    \phantom{de}\mbox{\textbf{def} $\delta(i,j)$: return \#columns differing in rows $i$,$j$ of $E$}\\
    \phantom{de}\textbf{def} cost = $\sum_{\mathrm{rows}\: i<j} |\delta(i,j)-d|$\\
  
    \begin{algorithmic} 
        \STATE \textbf{Initialize} $E$ with random permutation in each row
        \WHILE{cost $>0$}
        \STATE $r$ = random choice from \{rows $i$\}
        \STATE $c,d$ = randomly selected distinct columns in $r$
        \STATE $\Delta$ = (cost if $c$ and $d$ are swapped) - (current cost)
        \STATE \textbf{if} $\Delta<0$, or otherwise with probability $e^{-\Delta/0.444444}$:
        \STATE\hspace{\algorithmicindent} \textbf{swap} columns $c$ and $d$ in $r$
        \ENDWHILE
        \STATE \textbf{return} $E$
    \end{algorithmic}
\end{algorithm}

Broadly speaking, the application of genetic algorithms and simulated annealing to generating combinatorial designs is well established (\textit{Handbook} chapter VII.6).  However, from reviewing the literature, the application of genetic algorithms and simulated annealing (or local search) specifically to Symmetric and Skew Weighing Matrices, Balanced Ternary Designs, and Equidistant Permutation Arrays appears to be novel.  Genetic algorithms have been proposed for equidistant permutation arrays, but with no results reported yet \cite{epaga}.  

For Packing Arrays, some of the previous results from the literature were obtained in 2001 via genetic algorithms and simulated annealing \cite{stardom}.  Unlike the constant-temperature simulated annealing generated here, this earlier simulated annealing used a traditional cooling schedule that ramps temperature down towards 0 rather quickly.  Given increases in computing power since 2001, simulated annealing can now run for orders of magnitude more iterations, making constant temperature potentially more valuable.

For Florentine Rectangles, the successful generated code implements a depth-first search that randomly shuffles the choices to be tried at each position in the array.  This randomization enables effective use of the multiple random seeds that run in parallel under CPro1.  The code also achieves efficiency by incrementally tracking the status of constraints.  Even so, the code does not appear to be fully optimized (for example, it does not leverage bit parallelism).  A recent application of Florentine Rectangles to coding theory \cite{florentinecodes} presented new systematic constructions for Florentine Rectangles that scale up to arbitrarily large $n$, but noted that these constructions did not improve upon results from the \textit{Handbook} for small $n$.

All of the successful strategies are randomized.  This gives the possibility of running them further with different random seeds, to generate additional examples of the generated designs with the same parameters.

\subsection{Modes of Failure and Success}
\label{sec:failuremodes}

A majority of generated candidates fail completely and score 0 on the development instances.  Even for candidates which score more than 0, many are naive and have no hope of solving instances beyond the very smallest development instances.  These failures happen for a variety of reasons depending on the model.  For example, considering simulated annealing candidates for Equidistant Permutation Arrays (see Table~\ref{tab:models}):
\begin{description}
    \item[Claude 3.5 Sonnet's failed run] Here, 37 of the 50 simulated annealing candidates have an off-by-one bug when checking argc (the number of command-line arguments), stopping execution and scoring 0.  The generated code is required to take the instance-defining parameters, a random seed, and any hyperparameters as command-line arguments.  In C, the number of command-line arguments includes an extra one for the command itself; Claude 3.5 Sonnet's off-by-one error fails to account for this.  
    \item[Mistral Large's failed run] A majority of simulated annealing candidates either (a) have a bug that swaps between rows, destroying the property that each row is a permutation, or (b) drive temperature rapidly to 0, becoming stuck in local optima with no possibility of escape -- these candidates can score more than 0, but never solve more than a quarter of the development instances.  Only 3 of Mistral Large's simulated annealing candidates avoid problems like these and solve over a quarter of the development instances, but 3 candidates didn't provide enough experiments to optimize simulated annealing for the problem.
    \item[GPT-4o's successful run] A majority of its candidates which solve less than a quarter of the development instances either (a) drive temperature rapidly to 0 as with Mistral Large, or else (b) have a syntax error that prevents the C code from compiling.  But problems occur less frequently than with Mistral Large, and GPT-4o has 12 simulated annealing candidates that are viable experiments solving over a quarter of the development instances; enough experimentation that GPT-4o is able to succeed with a candidate that solves an open instance.  An example of such experimentation is with variations in the cost function.  Comparing to the eventual successful approach in Algorithm~\ref{alg:epacpro1}, we see variants which square the $|\delta(i,j)-d|$, or which only count the rows in which $|\delta(i,j)-d|>0$ instead of using the actual value of $|\delta(i,j)-d|$.  Such variants are reasonable experiments, but ultimately they have mediocre scores and likely would not be viable for solving open instances.  The cost function in Algorithm~\ref{alg:epacpro1}, when combined with other appropriate details (e.g. neighborhood) is able to succeed in solving an open instance. 
\end{description}

\section{Limitations}

The positive results reported here arise from an automation of computational experimentation that could have been done manually.  The research community has only undertaken limited effort on such computational experimentation for the designs with positive results here; this may have left low hanging fruit for CPro1.  Some of the designs with no positive results here have received greater attention from the research community.  For example, for Coverings, an online repository notes results from various contributors, who have used simulated annealing, local search, and other methods \cite{lajollacovering}.  In CPro1's run on Coverings, generated code was successful on development instances, but had no success on open instances.

The combinatorial design research community has greater focus on systematic mathematical constructions (e.g. using algebraic methods), rather than direct computational search for small designs.  Systematic constructions may be scaled up to much larger instances than the ones within reach of direct search by heuristics.  Even for small instances, some of the literature results in Tables~\ref{tab:mainresults}~and~\ref{tab:prototpyingresults} come from systematic mathematical constructions \cite{designs2002,georgiou}.   CPro1 doesn't necessarily exclude systematic constructions -- for example, for Costas Arrays, some of the generated code implements the Welch method of systematic construction \cite{costasmethods}, and this successfully solves larger development instances than are solved by direct search.  However, CPro1 is never able to extend these methods in a way that could succeed on open instances.

The positive results result from applying standard methods, not inventing new methods.  The protocol helps automate substantial experimentation in trying various standard methods, and testing variants (e.g. cost function variants as noted in Section~\ref{sec:failuremodes}) to optimize the method to the problem at hand -- but it is not inventing new techniques.

The results presented here are from two full runs of CPro1 on each type of combinatorial design -- one with GPT4o and one with Mistral Large.  Repeated runs would yield different results.  Since the LLMs are inherently nondeterministic in their responses, it isn't possible to exactly the replicate the results of any one run.

We did not test CPro1 at larger scale (e.g. 10,000 candidates per run rather than 1000), and it is possible this could give better results.

A test run of CPro1 on the Cap Set problem failed to reproduce FunSearch's result \cite{funsearch} of a size-512 cap set for dimension $n=8$.

\section{Code Availability}
The Python code for CPro1, along with the solutions to open instances and the generated C code that constructed them, are available: https://github.com/Constructive-Codes/CPro1
    
\section{Conclusion}

The protocol CPro1 uses LLMs to generate code, and has successfully solved open instances of the existence problem for 6 types of combinatorial designs. The protocol can be run on additional types of combinatorial designs, by supplying a textual definition, a Python verifier, and small collections of development instances and open instances.  The protocol could likely be applied in other domains that allow automated full verification of solutions.  The protocol can be readily used with new LLMs as they become available (e.g. ``reasoning models" like OpenAI's o1 \cite{openaio1}), which may be able to reduce failure modes and improve overall capabilities. 

\section*{Acknowledgements} Thanks to Mark Land for discussions and comments.

\bibliographystyle{named}
\bibliography{protocol}
\clearpage

\begin{algorithm*}[h!]
    \caption{Protocol CPro1.  Note \textbf{prompt}(x) returns LLM result when prompted with x.  See Figs.~\ref{fig:strategiesprompt}~to~\ref{fig:optprompt} for full prompts.}
    \label{alg:cpro1}
    \textbf{Input}: Definition (combinatorial design problem definition), Dev \& Open (instance parameter lists), Verifier (Python code) \\
    \textbf{Parameters for Code Generation}: N = 20 strategies per rep, R = 50 reps \\
    \textbf{Parameters for Execution}: Devseeds \& Openseeds (\# rand seeds, set so \#seeds * \# instances = 32 threads to run in parallel),\\
    \phantom{xi}Fulldevtime = 2 hours, Opentime = 48 hours (full run time limits) \\
    \textbf{Parameters for Hyperparameter Tuning}: Init\_gridsize = 1000 hyperparms, Init\_runtime = 0.5 seconds, Scale = 10\\
    \textbf{Parameters for Optimization}: Opt\_cand = 50, Opt\_rounds = 5, Opt\_delta = 0.1 \\
    \textbf{Parameters for Selection}: Devcand = 5 (\#candidates for opt. and final Dev testing), Opencand = 2 (\#cand. for Open instances)\\
    \textbf{Output}: Verified designs for the Open instances (if found)\\
    \\
    // Parallel sandboxed execution of a candidate executable E: \\
    \textbf{def} exec(E,Hyperparms,Inst,Seeds,Maxtime): \textbf{return} output \& runtime of E with Hyperparms for Maxtime, on \#Seeds $\times$ Inst\\[6pt]
    // Number of verified correct results, plus speed bonus in the range of $[0,1]$: \\
    \textbf{def} scoring(Results,Maxtime): \textbf{return} $\sum_{r \in \mathrm{Results}} [1 + (r$'s time/(Maxtime*$|$Results$|))]$ \textbf{if} $r$ passes Verifier check \\[6pt]
    // Hyperparameter tuning functions:\\
    \textbf{def} one\_grid(Min,Max,Points): \textbf{return} coarse linear interpolation in middle, plus logarithmic/fine-grained ends to Min\&Max\\[6pt]
    \textbf{def} hyperparm\_grid(Ranges): // Ranges has Min,Max,Default for each hyperparameter.\\
    \phantom{xi} // Allocate Init\_gridsize points equally across $|Ranges|+1$ grids as follows: \\
    \phantom{xi} $G_{\mathrm{balanced}}$ = cross product of one\_grid's for each hyperparameter \\
    \phantom{xi} $G_{i}$ = cross product of one\_grid for hyperparameter i with 3 values [Min,Max,Default] for each other hyperparameter \\
    \phantom{xi} \textbf{return} union of $G_{\mathrm{balanced}}$ and all $G_{i}$\\[6pt]
    \textbf{def} run\_grid(E,Grid,Time): \textbf{return} [Parms \& scoring(exec(E,Parms,Dev inst.,Devseeds,Time),Time) \textbf{for} Parms in Grid] \\[6pt]
    \textbf{def} hyper\_tune(E,Grid,Time): \\
    \phantom{xi} Results = run\_grid(E,Grid,Time)\\
    \phantom{xi} \textbf{if} $|$Grid$| \leq $ Scale: \textbf{return} single best-scoring Parms in Results\\
    \phantom{xi} \textbf{else}: \textbf{return} hyper\_tune(E, $(|$Grid$|/$Scale$)$ best-scoring Parms in Results, Time*Scale) // recurse: smaller grid, more time \\
    \phantom{xi} // Final tests in hyperparm\_tune always run for Init\_runtime*Scale$^{(\log_{\mathrm{Scale}} \mathrm{Init\_gridsize})-1}$ = 50 seconds\\
    \\
    // Main protocol:
    \begin{algorithmic} 
        \STATE \textbf{initialize} Candidates to empty array
        \FOR{R(=50) reps}
        \STATE Strategies = \textbf{prompt}(Definition + "Please suggest " + N(=20) + " different approaches we could implement in C...") 
        \FOR{S in Strategies}
        \STATE Details = \textbf{prompt}(Definition + "We have selected..." + S + "...Describe the elements of this approach...") 
        \STATE Code,Hyperparm\_ranges = \textbf{prompt}("Now implement this approach in C...") // Continue chat, so Details are in context.
        \STATE \textbf{append} Code,Hyperparm\_ranges to Candidates
        \ENDFOR
        \ENDFOR \phantom{x} // We now have R*N=1000 candidates\vspace{6pt}
        \FOR{C in Candidates}
        \STATE \textbf{Compile} C's source code to Executable
        \STATE \textbf{Set} C's Hyperparm\_settings,Score = hyper\_tune(Executable,hyperparm\_grid(C's Hyperparm\_ranges),Init\_runtime)
        \ENDFOR
        \STATE \textbf{truncate} Candidates to the best Devcand according to their Score\vspace{6pt}
        \FOR{C in Candidates}
        \FOR{i=1 to Opt\_rounds(=5)}
        \STATE O = [\textbf{prompt}("...make one small change to significantly improve the performance..." + C + "...") * Opt\_cand(=50) times] \\
        \STATE C' = highest-scoring single result of compiling, exec(), and scoring() each of the Opt\_cand candidates
        \STATE \textbf{if} C' scores better than C by at least Opt\_delta: \textbf{replace} C by C', \textbf{else}: break out of optimization for this C\\
        \ENDFOR
        \ENDFOR\vspace{6pt}
        \STATE \textbf{for} C in Candidates \textbf{do}: Devscore = scoring(exec(C's Executable,C's Hyperparm\_settings,Dev inst,Devseeds,Fulldevtime))
        \STATE \textbf{truncate} Candidates to the best Opencand according to their Devscore\vspace{6pt}
        \FOR{C in Candidates}
        \STATE Results = exec(C's Executable,C's Hyperparm\_settings,Open instances,Openseeds,Opentime)
        \STATE \textbf{output} each design in Results that passes Verifier check
        \ENDFOR        
        
    \end{algorithmic}
\end{algorithm*}

\appendix
\section{Protocol Pseudocode and Prompts}
\label{sec:pseudocode}

The protocol CPro1 that is outlined in Fig.~\ref{fig:protocol} is specified in more detail in Algorithm~\ref{alg:cpro1}.  The prompts that are used are shown in Figures~\ref{fig:strategiesprompt}~to~\ref{fig:optprompt}.  These are shown instantiated for the Packing Array (PA) combinatorial design problem; the text that can vary between problems is shown in italics and the rest of the prompt text is constant. 

\phantom{paragraph for alignment}
\\[9pt]

\begin{figure}[th!]
\centering
\fbox{\parbox{\linewidth}{
\textit{A "Packing Array" PA(N,k,v) is an N x k array (N rows and k columns), with each entry from the v-set \{0,1,...v-1\}, so that every N x 2 subarray contains every ordered pair of symbols at most once.  Given (N,k,v), we want to construct PA(N,k,v).} Please suggest 20 different approaches we could implement in C.  For now, just describe the approaches. Then I will pick one of the approaches, and you will write the C code to test it. We will start testing on small parameters like \textit{N=4 k=6 v=3}, and then once those work we will proceed to larger parameters like \textit{N=32 k=5 v=6}. Format your list items like this example: "12. **Strategy Name**: Sentences describing strategy, all on one line..."}}
\caption{Strategies prompt to the LLM, instantiated (italicized text) for Packing Arrays.}
\label{fig:strategiesprompt}
\end{figure}

\begin{figure}[th!]
\fbox{\parbox{\linewidth}{
\textit{A "Packing Array" PA(N,k,v) is an N x k array (N rows and k columns), with each entry from the v-set \{0,1,...v-1\}, so that every N x 2 subarray contains every ordered pair of symbols at most once.  Given (N,k,v), we want to construct PA(N,k,v).}\\
\\
We have selected the following approach:\\
\textit{Simulated Annealing: Use simulated annealing to slowly improve a randomly initialized array by making small changes and accepting them based on a cooling schedule.} Do not terminate until a valid solution is found.\\
\\
Describe the elements of this approach to constructing a \textit{Packing Array}.  Do not yet write code; just describe the details of the approach.}}
\caption{Details prompt to the LLM, instantiated (italicized text) for Packing Arrays with a Simulated Annealing strategy resulting from the prompt in Fig.~\ref{fig:strategiesprompt}.  Note "Do not terminate until a valid solution is found" is not italicized and is a fixed part of the template that is always appended to the strategy description generated by the LLM.}
\label{fig:detailsprompt}
\end{figure}

\begin{figure}[th!]
\fbox{\parbox{\linewidth}{
Now implement this approach in C, following in detail the plan described above.  Provide the complete code.  The code should only print out the final \textit{Packing Array} once a valid solution is found.\\
\\
I will be running the code from the Linux command line.  Please have the C code take command-line parameters: \textit{N k v} seed (in that order), followed by additional parameters as needed which represent hyperparameters of your approach.  The seed is the random seed (if no random seed is needed, still accept this parameter but ignore it).\\
\\
After giving the complete code, for each hyperparameter that is an extra command-line parameter, provide a specification in JSON with fields "name","min","max","default" specifying the name, minimum value, maximum value, and default value for the hyperparameter.  For example: {"name":"gamma", "min":0.0, "max":2.0, "default":0.5}.  If no hyperparameters are required then just state "No Hyperparameters Required" after giving the complete code.  I will be using Linux timeout to set a time limit on execution of your program, and for challenging \textit{PA} parameters this will be a long timeout (hours).  So to maximize chances of finding a solution your code should keep running indefinitely until it finds a valid solution.  Therefore, eliminate hyperparameters that would control termination, since your program needn't terminate until it succeeds.\\
\\
We will start testing with small problem parameters like \textit{N=4 k=6 v=3}.  Once those work, we can then test further refinements and move towards larger problem parameters like \textit{N=32 k=5 v=6}.}}
\caption{Code generation prompt to the LLM, instantiated (italicized text) for Packing Arrays.  Note this continues the chat following on from the Details prompt in Fig.~\ref{fig:detailsprompt}; so the Details prompt and response are in context when the LLM responds to this prompt.}
\label{fig:codeprompt}
\end{figure}

\begin{figure*}[t!]
\fbox{\parbox{\linewidth}{
\textit{A "Packing Array" PA(N,k,v) is an N x k array (N rows and k columns), with each entry from the v-set \{0,1,...v-1\}, so that every N x 2 subarray contains every ordered pair of symbols at most once.  Given (N,k,v), we want to construct PA(N,k,v).}\\
\\
We have selected the following approach:\\
\textit{Simulated Annealing: Use simulated annealing to slowly improve a randomly initialized array by making small changes and accepting them based on a cooling schedule.} Do not terminate until a valid solution is found.\\
\\
The C code below implements this approach.  It takes command-line parameters: \textit{N k v} seed (in that order) where seed is the random number generator seed, followed by additional parameters which represent hyperparameters of the approach.\\
\\
We are seeking to make one small change to significantly improve the performance of this code.\\
\\
\textit{\#include $<$stdio.h$>$\\
\#include $<$stdlib.h$>$\\
...}\\
\\
With these hyperparameters chosen by hyperparameter tuning: \textit{T0=0.1575068233468584, cooling\_rate=1.0}.  This code solves \textit{N=4 k=6 v=3} in an average of \textit{0.0026} seconds across \textit{2} attempts, solves \textit{N=6 k=5 v=3} in an average of \textit{0.0028} seconds across \textit{2} attempts, solves \textit{N=9 k=6 v=4} in an average of \textit{0.0686} seconds across \textit{2} attempts, solves \textit{N=20 k=6 v=5} in an average of \textit{0.0789} seconds across \textit{2} attempts, solves \textit{N=25 k=6 v=5} in an average of \textit{0.0306} seconds across \textit{2} attempts, solves \textit{N=25 k=5 v=6} in an average of \textit{0.0139} seconds across \textit{2} attempts, solves \textit{N=30 k=5 v=6} in an average of \textit{0.9323} seconds across \textit{2} attempts, solves \textit{N=31 k=5 v=6} in \textit{1} out of \textit{2} attempts with a time limit of 50.0 seconds each, solves \textit{N=16 k=7 v=6} in an average of \textit{0.011} seconds across \textit{2} attempts, solves \textit{N=23 k=7 v=6} in \textit{1} out of \textit{2} attempts with a time limit of 50.0 seconds each, solves \textit{N=17 k=10 v=8} in an average of \textit{0.014} seconds across \textit{2} attempts, solves \textit{N=22 k=10 v=8} in an average of \textit{0.1002} seconds across \textit{2} attempts, solves \textit{N=25 k=10 v=8} in an average of \textit{6.9132} seconds across \textit{2} attempts, solves \textit{N=14 k=11 v=8} in an average of \textit{0.0095} seconds across \textit{2} attempts, solves \textit{N=19 k=11 v=8} in an average of \textit{0.0396} seconds across \textit{2} attempts, and solves \textit{N=22 k=11 v=8} in an average of \textit{4.2044} seconds across \textit{2} attempts.  We want to make one small change to the code to significantly improve performance, without parallelizing or changing the algorithm or the hyperparameter handling.  Describe your plan for making one small change to significantly improve performance.  Then, implement your plan and provide the complete updated code.}}
\caption{Optimization prompt to the LLM for generating candidate optimized code, instantiated (italicized text) for a Packing Arrays run with a Simulated Annealing strategy resulting from the prompt in Fig.~\ref{fig:strategiesprompt}, code (shown in abbreviated form) resulting from the prompt in Fig.~\ref{fig:codeprompt}, and results from automated hyperparameter tuning.}
\label{fig:optprompt}
\end{figure*}

\clearpage
\section{Solutions to Open Instances of Combinatorial Design Problems}
\label{sec:solutions}

Figures~\ref{fig:arrays1}~to~\ref{fig:arrays22} show the verified designs that resolve open instances of combinatorial design problems as noted in Tables~\ref{tab:mainresults}~and~\ref{tab:prototpyingresults}.  Each of these was constructed by code that was generated by protocol CPro1.

\begin{figure}[h!]
\centering
\[
\fbox{
$\begin{array}{*{7}{r}}
1 & 3 & 2 & 5 & 5 & 2 & 1\\ 
5 & 1 & 3 & 1 & 0 & 5 & 1\\ 
4 & 2 & 0 & 1 & 2 & 2 & 4\\ 
0 & 2 & 1 & 4 & 5 & 5 & 0\\ 
2 & 4 & 0 & 5 & 1 & 1 & 0\\ 
2 & 5 & 5 & 2 & 0 & 2 & 5\\ 
5 & 0 & 1 & 3 & 1 & 2 & 2\\ 
1 & 5 & 3 & 3 & 2 & 0 & 0\\ 
3 & 4 & 3 & 2 & 5 & 3 & 4\\ 
4 & 1 & 1 & 2 & 3 & 1 & 3\\ 
2 & 2 & 2 & 3 & 4 & 3 & 3\\ 
1 & 0 & 5 & 0 & 3 & 5 & 4\\ 
4 & 4 & 2 & 0 & 0 & 0 & 2\\ 
4 & 5 & 4 & 4 & 1 & 3 & 1\\ 
2 & 3 & 3 & 4 & 3 & 4 & 2\\ 
1 & 4 & 1 & 1 & 4 & 4 & 5\\ 
3 & 1 & 4 & 0 & 4 & 2 & 0\\ 
3 & 3 & 5 & 1 & 1 & 0 & 3\\ 
5 & 5 & 0 & 0 & 5 & 4 & 3\\ 
0 & 1 & 5 & 5 & 2 & 3 & 2\\ 
3 & 0 & 2 & 4 & 2 & 1 & 5\\ 
0 & 3 & 4 & 3 & 0 & 1 & 4\\ 
5 & 2 & 4 & 5 & 3 & 0 & 5\\ 
0 & 0 & 0 & 2 & 4 & 0 & 1

\end{array}$}
\]
\caption{Packing Array with N=24 k=7 v=6}
\label{fig:arrays4}
\end{figure}

\begin{figure}[h!]
\centering
\[
\fbox{
$\begin{array}{*{8}{r}}
3 & 4 & 2 & 3 & 4 & 4 & 0 & 5\\ 
4 & 1 & 1 & 5 & 4 & 0 & 1 & 3\\ 
3 & 1 & 4 & 0 & 1 & 3 & 4 & 1\\ 
0 & 5 & 2 & 1 & 5 & 0 & 4 & 2\\ 
4 & 4 & 5 & 1 & 3 & 3 & 5 & 0\\ 
4 & 2 & 3 & 4 & 5 & 5 & 0 & 1\\ 
2 & 0 & 1 & 2 & 5 & 3 & 2 & 5\\ 
3 & 3 & 3 & 2 & 3 & 2 & 3 & 3\\ 
1 & 3 & 5 & 5 & 0 & 5 & 4 & 5\\ 
2 & 3 & 2 & 4 & 1 & 1 & 1 & 0\\ 
1 & 0 & 3 & 3 & 1 & 0 & 5 & 4\\ 
1 & 1 & 0 & 4 & 3 & 4 & 2 & 2\\ 
5 & 5 & 1 & 0 & 3 & 1 & 0 & 4\\ 
2 & 2 & 4 & 1 & 0 & 4 & 3 & 4\\ 
0 & 2 & 5 & 3 & 2 & 1 & 2 & 3\\ 
5 & 4 & 4 & 2 & 2 & 5 & 1 & 2\\ 
2 & 5 & 0 & 5 & 2 & 2 & 5 & 1\\ 
0 & 0 & 0 & 0 & 4 & 5 & 3 & 0
\end{array}$}
\]
\caption{Packing Array with N=18 k=8 v=6}
\label{fig:arrays5}
\end{figure}

\begin{figure}[h!]
\centering
\[
\fbox{
$\begin{array}{*{10}{r}}
2 & 5 & 0 & 2 & 2 & 1 & 0 & 0 & 1 & 5\\ 
3 & 3 & 0 & 7 & 6 & 7 & 1 & 4 & 2 & 3\\ 
5 & 4 & 2 & 1 & 0 & 7 & 3 & 0 & 7 & 2\\ 
0 & 4 & 5 & 2 & 5 & 6 & 2 & 4 & 0 & 7\\ 
6 & 3 & 5 & 0 & 2 & 0 & 7 & 5 & 3 & 2\\ 
6 & 4 & 0 & 6 & 4 & 3 & 4 & 6 & 6 & 6\\ 
7 & 6 & 1 & 3 & 2 & 7 & 5 & 1 & 0 & 6\\ 
3 & 7 & 3 & 0 & 5 & 4 & 5 & 6 & 1 & 0\\ 
5 & 6 & 3 & 7 & 7 & 2 & 7 & 2 & 6 & 7\\ 
0 & 3 & 2 & 4 & 1 & 2 & 6 & 3 & 1 & 6\\ 
5 & 2 & 6 & 5 & 1 & 1 & 5 & 4 & 3 & 1\\ 
4 & 0 & 4 & 7 & 1 & 5 & 0 & 6 & 0 & 2\\ 
0 & 2 & 1 & 0 & 6 & 3 & 0 & 2 & 7 & 4\\ 
1 & 1 & 0 & 3 & 1 & 0 & 2 & 2 & 5 & 0\\ 
4 & 2 & 2 & 6 & 2 & 6 & 1 & 7 & 4 & 0\\ 
7 & 5 & 5 & 5 & 6 & 5 & 3 & 3 & 6 & 0\\ 
4 & 1 & 1 & 5 & 4 & 4 & 6 & 0 & 2 & 7\\ 
6 & 2 & 3 & 2 & 3 & 5 & 6 & 1 & 5 & 3\\ 
4 & 5 & 6 & 3 & 5 & 2 & 4 & 5 & 7 & 3\\ 
7 & 3 & 4 & 2 & 0 & 4 & 4 & 2 & 4 & 1\\ 
6 & 6 & 4 & 4 & 5 & 1 & 3 & 7 & 2 & 4\\ 
7 & 1 & 7 & 4 & 3 & 6 & 7 & 6 & 7 & 5\\ 
2 & 7 & 4 & 3 & 3 & 3 & 1 & 3 & 3 & 7\\ 
1 & 6 & 7 & 1 & 4 & 5 & 1 & 5 & 1 & 1\\ 
0 & 7 & 6 & 7 & 4 & 0 & 3 & 1 & 4 & 5\\ 
3 & 0 & 7 & 6 & 0 & 2 & 2 & 1 & 3 & 4\\ 
2 & 0 & 6 & 1 & 6 & 4 & 7 & 7 & 5 & 6\\ 
1 & 0 & 5 & 4 & 7 & 3 & 5 & 0 & 4 & 3

\end{array}$}
\]
\caption{Packing Array with N=28 k=10 v=8}
\label{fig:arrays6}
\end{figure}

\begin{figure}[h!]
\centering
\[
\fbox{
$\begin{array}{*{11}{r}}
7 & 3 & 3 & 2 & 4 & 4 & 1 & 0 & 2 & 7 & 3\\ 
2 & 5 & 1 & 6 & 7 & 2 & 7 & 0 & 4 & 6 & 6\\ 
7 & 1 & 5 & 0 & 7 & 6 & 0 & 5 & 6 & 4 & 5\\ 
6 & 2 & 4 & 5 & 5 & 4 & 2 & 4 & 3 & 6 & 5\\ 
6 & 0 & 3 & 0 & 2 & 0 & 5 & 2 & 5 & 0 & 6\\ 
0 & 3 & 4 & 6 & 3 & 3 & 4 & 2 & 6 & 2 & 7\\ 
7 & 7 & 2 & 6 & 6 & 7 & 3 & 4 & 1 & 0 & 4\\ 
1 & 1 & 2 & 3 & 5 & 3 & 7 & 1 & 2 & 1 & 1\\ 
5 & 4 & 2 & 4 & 1 & 0 & 0 & 3 & 4 & 2 & 3\\ 
5 & 5 & 6 & 1 & 5 & 5 & 3 & 2 & 0 & 7 & 0\\ 
3 & 1 & 4 & 1 & 0 & 1 & 1 & 7 & 4 & 0 & 2\\ 
1 & 0 & 0 & 7 & 3 & 2 & 3 & 7 & 3 & 4 & 3\\ 
4 & 7 & 6 & 0 & 1 & 1 & 4 & 0 & 3 & 3 & 1\\ 
3 & 4 & 6 & 5 & 4 & 2 & 5 & 6 & 6 & 1 & 4\\ 
4 & 6 & 0 & 5 & 7 & 3 & 6 & 3 & 7 & 0 & 0\\ 
1 & 6 & 5 & 2 & 2 & 5 & 4 & 6 & 1 & 6 & 2\\ 
3 & 6 & 3 & 7 & 1 & 6 & 7 & 4 & 0 & 5 & 7\\ 
0 & 0 & 7 & 4 & 0 & 6 & 2 & 0 & 1 & 1 & 0\\ 
4 & 2 & 1 & 4 & 6 & 5 & 1 & 1 & 5 & 4 & 7\\ 
0 & 4 & 5 & 3 & 6 & 4 & 6 & 7 & 0 & 3 & 6\\ 
2 & 7 & 7 & 1 & 3 & 0 & 6 & 6 & 2 & 5 & 5\\ 
5 & 3 & 1 & 3 & 2 & 1 & 2 & 5 & 7 & 5 & 4\\ 
2 & 2 & 0 & 2 & 0 & 7 & 5 & 5 & 0 & 2 & 1\\ 
6 & 5 & 7 & 7 & 4 & 7 & 0 & 1 & 7 & 3 & 2

\end{array}$}
\]
\caption{Packing Array with N=24 k=11 v=8}
\label{fig:arrays7}
\end{figure}

\begin{figure}[h!]
\centering
\[
\fbox{
$\begin{array}{*{11}{r}}
6 & 4 & 2 & 7 & 8 & 5 & 6 & 8 & 8 & 3 & 0\\ 
8 & 2 & 5 & 5 & 3 & 1 & 5 & 6 & 0 & 3 & 2\\ 
1 & 1 & 6 & 1 & 0 & 1 & 0 & 7 & 7 & 2 & 0\\ 
3 & 2 & 8 & 4 & 8 & 2 & 0 & 4 & 1 & 4 & 1\\ 
3 & 5 & 0 & 7 & 5 & 0 & 1 & 6 & 2 & 2 & 6\\ 
0 & 1 & 1 & 4 & 5 & 7 & 5 & 8 & 6 & 1 & 5\\ 
1 & 7 & 7 & 7 & 6 & 8 & 3 & 2 & 1 & 8 & 5\\ 
7 & 8 & 5 & 6 & 0 & 2 & 7 & 2 & 8 & 1 & 4\\ 
2 & 3 & 3 & 3 & 3 & 0 & 0 & 1 & 6 & 8 & 4\\ 
5 & 6 & 5 & 0 & 2 & 7 & 1 & 1 & 1 & 6 & 0\\ 
5 & 0 & 2 & 8 & 0 & 8 & 2 & 6 & 6 & 4 & 7\\ 
3 & 1 & 7 & 6 & 3 & 5 & 2 & 3 & 3 & 6 & 3\\ 
1 & 4 & 3 & 0 & 5 & 2 & 8 & 5 & 3 & 0 & 7\\ 
4 & 8 & 2 & 4 & 2 & 1 & 4 & 0 & 2 & 8 & 3\\ 
1 & 2 & 2 & 3 & 7 & 7 & 7 & 3 & 5 & 5 & 6\\ 
5 & 4 & 6 & 5 & 6 & 0 & 4 & 3 & 4 & 1 & 1\\ 
6 & 3 & 6 & 2 & 5 & 6 & 2 & 0 & 1 & 5 & 2\\ 
7 & 3 & 8 & 0 & 1 & 5 & 4 & 6 & 7 & 7 & 5\\ 
0 & 7 & 4 & 3 & 0 & 5 & 1 & 0 & 0 & 0 & 1\\ 
6 & 5 & 4 & 8 & 6 & 1 & 7 & 1 & 3 & 7 & 8\\ 
6 & 2 & 1 & 6 & 1 & 3 & 1 & 7 & 4 & 8 & 7\\ 
4 & 6 & 0 & 6 & 6 & 4 & 0 & 8 & 5 & 0 & 2\\ 
0 & 3 & 5 & 8 & 8 & 4 & 3 & 5 & 4 & 2 & 3\\ 
8 & 0 & 6 & 3 & 1 & 2 & 3 & 8 & 2 & 6 & 8\\ 
2 & 0 & 0 & 1 & 2 & 3 & 5 & 2 & 3 & 5 & 1\\ 
4 & 0 & 7 & 2 & 8 & 7 & 8 & 7 & 0 & 7 & 4\\ 
8 & 8 & 8 & 2 & 4 & 8 & 6 & 1 & 4 & 0 & 6\\ 
7 & 1 & 4 & 2 & 2 & 0 & 3 & 4 & 5 & 3 & 7\\ 
2 & 7 & 1 & 5 & 4 & 6 & 7 & 5 & 2 & 4 & 0\\ 
5 & 8 & 1 & 7 & 3 & 4 & 8 & 4 & 7 & 5 & 8\\ 
7 & 6 & 7 & 5 & 7 & 3 & 6 & 0 & 6 & 2 & 8\\ 
0 & 5 & 3 & 1 & 7 & 8 & 4 & 4 & 8 & 6 & 2

\end{array}$}
\]
\caption{Packing Array with N=32 k=11 v=9}
\label{fig:arrays8}
\end{figure}

\begin{figure}[h!]
\centering
\[
\fbox{
$\begin{array}{*{12}{r}}
3 & 2 & 4 & 8 & 1 & 8 & 8 & 1 & 0 & 8 & 5 & 0\\ 
3 & 1 & 2 & 7 & 0 & 3 & 5 & 8 & 1 & 2 & 2 & 7\\ 
1 & 6 & 5 & 7 & 5 & 7 & 4 & 1 & 8 & 5 & 7 & 8\\ 
2 & 7 & 8 & 0 & 7 & 5 & 7 & 5 & 0 & 5 & 0 & 2\\ 
7 & 3 & 8 & 3 & 5 & 1 & 6 & 0 & 2 & 7 & 5 & 7\\ 
3 & 0 & 8 & 1 & 4 & 6 & 4 & 2 & 3 & 0 & 6 & 3\\ 
2 & 5 & 6 & 4 & 0 & 0 & 0 & 0 & 5 & 6 & 6 & 0\\ 
8 & 2 & 7 & 0 & 3 & 0 & 3 & 7 & 1 & 0 & 7 & 6\\ 
6 & 8 & 2 & 5 & 3 & 8 & 2 & 5 & 4 & 7 & 6 & 4\\ 
0 & 5 & 3 & 0 & 4 & 4 & 8 & 3 & 8 & 7 & 2 & 5\\ 
0 & 0 & 1 & 4 & 6 & 2 & 6 & 1 & 1 & 1 & 0 & 4\\ 
7 & 7 & 4 & 4 & 3 & 7 & 1 & 6 & 3 & 2 & 1 & 5\\ 
0 & 2 & 6 & 3 & 2 & 6 & 1 & 8 & 4 & 5 & 3 & 1\\ 
5 & 3 & 5 & 5 & 4 & 5 & 0 & 6 & 1 & 8 & 8 & 1\\ 
4 & 6 & 4 & 2 & 2 & 5 & 3 & 0 & 6 & 4 & 2 & 4\\ 
7 & 8 & 0 & 7 & 7 & 0 & 8 & 4 & 6 & 1 & 4 & 1\\ 
4 & 8 & 6 & 6 & 1 & 7 & 7 & 3 & 7 & 0 & 8 & 7\\ 
0 & 6 & 0 & 8 & 8 & 3 & 7 & 2 & 2 & 6 & 1 & 6\\ 
2 & 1 & 5 & 3 & 6 & 8 & 3 & 2 & 7 & 3 & 4 & 5\\ 
1 & 3 & 7 & 1 & 2 & 4 & 2 & 4 & 7 & 2 & 0 & 0\\ 
6 & 5 & 7 & 6 & 6 & 1 & 4 & 8 & 6 & 8 & 1 & 2\\ 
5 & 7 & 2 & 6 & 8 & 6 & 6 & 7 & 8 & 4 & 4 & 0\\ 
6 & 4 & 1 & 2 & 5 & 6 & 5 & 4 & 0 & 6 & 8 & 5\\ 
8 & 8 & 8 & 8 & 6 & 4 & 5 & 6 & 5 & 4 & 3 & 8\\ 
8 & 4 & 4 & 7 & 8 & 1 & 0 & 3 & 4 & 3 & 0 & 3\\ 
1 & 1 & 0 & 2 & 1 & 2 & 0 & 7 & 3 & 7 & 3 & 2\\ 
4 & 4 & 3 & 1 & 0 & 2 & 1 & 5 & 2 & 8 & 4 & 8\\ 
5 & 0 & 3 & 2 & 7 & 7 & 2 & 8 & 5 & 3 & 5 & 6

\end{array}$}
\]
\caption{Packing Array with N=28 k=12 v=9}
\label{fig:arrays9}
\end{figure}

\begin{figure}[h!]
\centering
\[
\fbox{
$\begin{array}{*{14}{r}}
4 & 8 & 1 & 1 & 3 & 1 & 7 & 6 & 8 & 8 & 6 & 8 & 1 & 7\\ 
5 & 2 & 7 & 2 & 4 & 0 & 1 & 5 & 4 & 7 & 1 & 1 & 6 & 7\\ 
6 & 1 & 2 & 5 & 5 & 1 & 0 & 7 & 5 & 3 & 3 & 6 & 6 & 2\\ 
4 & 5 & 6 & 8 & 4 & 8 & 6 & 2 & 1 & 2 & 8 & 6 & 4 & 4\\ 
1 & 4 & 5 & 0 & 6 & 0 & 6 & 4 & 8 & 5 & 3 & 5 & 0 & 0\\ 
3 & 6 & 0 & 3 & 3 & 8 & 2 & 7 & 7 & 7 & 2 & 4 & 0 & 5\\ 
1 & 0 & 2 & 1 & 8 & 3 & 5 & 8 & 6 & 6 & 1 & 4 & 2 & 4\\ 
7 & 8 & 0 & 2 & 1 & 3 & 3 & 3 & 5 & 2 & 4 & 3 & 3 & 0\\ 
7 & 5 & 7 & 7 & 7 & 6 & 8 & 8 & 7 & 4 & 3 & 8 & 5 & 3\\ 
3 & 3 & 8 & 5 & 4 & 5 & 4 & 6 & 6 & 5 & 0 & 3 & 8 & 3\\ 
4 & 4 & 2 & 4 & 0 & 4 & 8 & 3 & 2 & 7 & 7 & 2 & 8 & 6\\ 
1 & 3 & 4 & 3 & 7 & 7 & 0 & 1 & 1 & 0 & 4 & 1 & 1 & 6\\ 
0 & 6 & 5 & 5 & 8 & 4 & 1 & 1 & 3 & 1 & 8 & 8 & 3 & 8\\ 
2 & 2 & 3 & 7 & 1 & 8 & 7 & 4 & 3 & 3 & 0 & 7 & 2 & 6\\ 
8 & 0 & 0 & 0 & 5 & 7 & 4 & 2 & 0 & 1 & 7 & 7 & 5 & 7\\ 
8 & 2 & 5 & 8 & 2 & 6 & 2 & 6 & 2 & 6 & 4 & 0 & 7 & 2\\ 
6 & 4 & 6 & 6 & 8 & 7 & 7 & 0 & 4 & 4 & 2 & 3 & 7 & 1\\ 
0 & 7 & 3 & 6 & 2 & 3 & 8 & 5 & 0 & 5 & 5 & 6 & 1 & 5\\ 
5 & 1 & 1 & 7 & 6 & 2 & 2 & 2 & 6 & 0 & 5 & 2 & 3 & 1\\ 
2 & 7 & 8 & 0 & 0 & 2 & 3 & 8 & 1 & 8 & 2 & 0 & 6 & 8\\ 
5 & 7 & 4 & 4 & 1 & 5 & 5 & 0 & 7 & 1 & 6 & 5 & 4 & 2

\end{array}$}
\]
\caption{Packing Array with N=21 k=14 v=9}
\label{fig:arrays10}
\end{figure}

\begin{figure}[h!]
\centering
\[
\fbox{
$\begin{array}{*{15}{r}}
8 & 4 & 1 & 2 & 2 & 5 & 5 & 0 & 5 & 8 & 3 & 7 & 0 & 4 & 7\\ 
3 & 8 & 6 & 1 & 4 & 6 & 6 & 2 & 7 & 3 & 8 & 8 & 5 & 6 & 7\\ 
5 & 8 & 2 & 5 & 5 & 1 & 4 & 5 & 4 & 0 & 4 & 5 & 0 & 1 & 2\\ 
3 & 5 & 4 & 5 & 3 & 4 & 3 & 3 & 3 & 1 & 1 & 7 & 2 & 2 & 3\\ 
0 & 3 & 8 & 7 & 4 & 7 & 7 & 0 & 6 & 2 & 5 & 3 & 2 & 1 & 5\\ 
7 & 0 & 3 & 4 & 3 & 7 & 5 & 5 & 2 & 3 & 6 & 6 & 1 & 0 & 1\\ 
8 & 6 & 3 & 3 & 5 & 3 & 2 & 1 & 3 & 4 & 7 & 3 & 5 & 8 & 4\\ 
7 & 1 & 8 & 6 & 1 & 5 & 3 & 6 & 7 & 4 & 2 & 1 & 6 & 5 & 2\\ 
5 & 4 & 4 & 7 & 8 & 0 & 1 & 6 & 1 & 7 & 8 & 0 & 7 & 8 & 1\\ 
1 & 7 & 8 & 2 & 0 & 1 & 8 & 8 & 0 & 5 & 8 & 2 & 1 & 2 & 4\\ 
6 & 1 & 1 & 1 & 3 & 8 & 2 & 8 & 4 & 7 & 5 & 4 & 4 & 7 & 0\\ 
4 & 3 & 0 & 4 & 6 & 4 & 4 & 1 & 0 & 6 & 3 & 8 & 7 & 5 & 0\\ 
2 & 3 & 6 & 3 & 0 & 2 & 0 & 3 & 8 & 8 & 2 & 0 & 4 & 0 & 6\\ 
2 & 6 & 5 & 0 & 6 & 8 & 6 & 0 & 1 & 0 & 1 & 2 & 6 & 3 & 8\\ 
1 & 1 & 0 & 8 & 8 & 3 & 7 & 7 & 8 & 1 & 4 & 6 & 8 & 3 & 7\\ 
4 & 7 & 7 & 5 & 2 & 0 & 0 & 2 & 2 & 4 & 0 & 4 & 3 & 3 & 5\\ 
8 & 5 & 5 & 8 & 1 & 6 & 8 & 4 & 6 & 6 & 6 & 5 & 3 & 7 & 6\\ 
0 & 0 & 2 & 6 & 7 & 4 & 2 & 4 & 5 & 5 & 0 & 0 & 8 & 6 & 8\\ 
6 & 2 & 7 & 0 & 7 & 2 & 1 & 7 & 7 & 2 & 7 & 5 & 1 & 4 & 3

\end{array}$}
\]
\caption{Packing Array with N=19 k=15 v=9}
\label{fig:arrays11}
\end{figure}

\begin{figure*}[h!]
\centering
\[
\fbox{
$\begin{array}{*{19}{r}}
0 & 0 & 0 & 0 & -1 & 0 & 1 & 1 & 0 & 0 & 1 & -1 & 0 & 1 & 1 & 0 & 1 & 1 & 0\\ 
0 & -1 & -1 & 1 & 0 & 0 & 0 & 0 & 1 & 1 & -1 & 0 & 0 & 0 & 1 & -1 & 0 & 0 & -1\\ 
0 & -1 & 0 & 0 & 1 & 0 & 0 & 1 & 0 & 1 & 1 & 1 & 0 & 1 & 0 & 1 & 0 & -1 & 0\\ 
0 & 1 & 0 & 0 & -1 & -1 & 0 & 1 & 0 & 1 & 0 & 0 & 0 & -1 & 0 & 1 & -1 & 0 & -1\\ 
-1 & 0 & 1 & -1 & 0 & -1 & 0 & -1 & 1 & 1 & 0 & 0 & 0 & 0 & 1 & 0 & 0 & 0 & 1\\ 
0 & 0 & 0 & -1 & -1 & 0 & 0 & 1 & 1 & -1 & 0 & 1 & -1 & 0 & 0 & -1 & 0 & -1 & 0\\ 
1 & 0 & 0 & 0 & 0 & 0 & -1 & 0 & 1 & -1 & 0 & -1 & 1 & 0 & 1 & 1 & 0 & -1 & 0\\ 
1 & 0 & 1 & 1 & -1 & 1 & 0 & 0 & 1 & 1 & 0 & 0 & 0 & 0 & -1 & 0 & 0 & 0 & 1\\ 
0 & 1 & 0 & 0 & 1 & 1 & 1 & 1 & 0 & 0 & -1 & 0 & 0 & 0 & 1 & 0 & -1 & 0 & 1\\ 
0 & 1 & 1 & 1 & 1 & -1 & -1 & 1 & 0 & 0 & 0 & 0 & 0 & 0 & 0 & -1 & 1 & 0 & 0\\ 
1 & -1 & 1 & 0 & 0 & 0 & 0 & 0 & -1 & 0 & 1 & 0 & 0 & -1 & 1 & -1 & -1 & 0 & 0\\ 
-1 & 0 & 1 & 0 & 0 & 1 & -1 & 0 & 0 & 0 & 0 & -1 & -1 & 1 & 0 & 0 & -1 & 0 & -1\\ 
0 & 0 & 0 & 0 & 0 & -1 & 1 & 0 & 0 & 0 & 0 & -1 & 1 & 1 & -1 & -1 & -1 & -1 & 0\\ 
1 & 0 & 1 & -1 & 0 & 0 & 0 & 0 & 0 & 0 & -1 & 1 & 1 & 1 & 0 & 0 & 0 & 1 & -1\\ 
1 & 1 & 0 & 0 & 1 & 0 & 1 & -1 & 1 & 0 & 1 & 0 & -1 & 0 & 0 & 0 & 0 & 0 & -1\\ 
0 & -1 & 1 & 1 & 0 & -1 & 1 & 0 & 0 & -1 & -1 & 0 & -1 & 0 & 0 & 1 & 0 & 0 & 0\\ 
1 & 0 & 0 & -1 & 0 & 0 & 0 & 0 & -1 & 1 & -1 & -1 & -1 & 0 & 0 & 0 & 1 & -1 & 0\\ 
1 & 0 & -1 & 0 & 0 & -1 & -1 & 0 & 0 & 0 & 0 & 0 & -1 & 1 & 0 & 0 & -1 & 1 & 1\\ 
0 & -1 & 0 & -1 & 1 & 0 & 0 & 1 & 1 & 0 & 0 & -1 & 0 & -1 & -1 & 0 & 0 & 1 & 0
\end{array}$}
\]
\caption{Symmetric Weighing Matrix with n=19 w=9}
\label{fig:arrays1}
\end{figure*}

\begin{figure*}[h!]
\centering
\[
\fbox{
$\begin{array}{*{21}{r}}
-1 & 0 & 0 & -1 & 1 & 0 & 0 & 0 & 0 & 1 & 0 & -1 & -1 & 0 & 0 & 0 & 0 & -1 & 1 & -1 & 0\\ 
0 & 0 & 1 & 1 & 0 & 0 & 0 & 1 & -1 & 0 & -1 & 0 & 0 & -1 & 1 & 0 & 0 & 0 & 1 & 0 & -1\\ 
0 & 1 & -1 & 0 & 0 & 0 & 0 & -1 & 1 & 0 & 0 & 0 & -1 & -1 & 1 & 0 & 0 & 1 & 0 & 0 & -1\\ 
-1 & 1 & 0 & 0 & 0 & 1 & 0 & 1 & 0 & 0 & 0 & 1 & 0 & 0 & 0 & -1 & -1 & 0 & -1 & -1 & 0\\ 
1 & 0 & 0 & 0 & 1 & 0 & 1 & 0 & 0 & 1 & 0 & 0 & 1 & 0 & 1 & 0 & 1 & 0 & -1 & -1 & 0\\ 
0 & 0 & 0 & 1 & 0 & 1 & -1 & 0 & 0 & 0 & 1 & 0 & 0 & -1 & -1 & 1 & 1 & 0 & 0 & -1 & 0\\ 
0 & 0 & 0 & 0 & 1 & -1 & 1 & 1 & 0 & 0 & 0 & 1 & -1 & 0 & -1 & 1 & 0 & 1 & 0 & 0 & 0\\ 
0 & 1 & -1 & 1 & 0 & 0 & 1 & 1 & 0 & 0 & 1 & -1 & 0 & 0 & 0 & 0 & 0 & -1 & 0 & 1 & 0\\ 
0 & -1 & 1 & 0 & 0 & 0 & 0 & 0 & 0 & 1 & 1 & 0 & -1 & -1 & 0 & -1 & 0 & 0 & -1 & 1 & 0\\ 
1 & 0 & 0 & 0 & 1 & 0 & 0 & 0 & 1 & -1 & -1 & 0 & 0 & -1 & -1 & -1 & 0 & -1 & 0 & 0 & 0\\ 
0 & -1 & 0 & 0 & 0 & 1 & 0 & 1 & 1 & -1 & 0 & 0 & -1 & 1 & 1 & 0 & 1 & 0 & 0 & 0 & 0\\ 
-1 & 0 & 0 & 1 & 0 & 0 & 1 & -1 & 0 & 0 & 0 & 1 & 0 & 0 & 0 & -1 & 1 & 0 & 1 & 0 & 1\\ 
-1 & 0 & -1 & 0 & 1 & 0 & -1 & 0 & -1 & 0 & -1 & 0 & 0 & 0 & 0 & 0 & 1 & 0 & -1 & 1 & 0\\ 
0 & -1 & -1 & 0 & 0 & -1 & 0 & 0 & -1 & -1 & 1 & 0 & 0 & 0 & 0 & -1 & 0 & 0 & 0 & -1 & -1\\ 
0 & 1 & 1 & 0 & 1 & -1 & -1 & 0 & 0 & -1 & 1 & 0 & 0 & 0 & 1 & 0 & 0 & 0 & 0 & 0 & 1\\ 
0 & 0 & 0 & -1 & 0 & 1 & 1 & 0 & -1 & -1 & 0 & -1 & 0 & -1 & 0 & 0 & 0 & 1 & 0 & 0 & 1\\ 
0 & 0 & 0 & -1 & 1 & 1 & 0 & 0 & 0 & 0 & 1 & 1 & 1 & 0 & 0 & 0 & 0 & 0 & 1 & 1 & -1\\ 
-1 & 0 & 1 & 0 & 0 & 0 & 1 & -1 & 0 & -1 & 0 & 0 & 0 & 0 & 0 & 1 & 0 & -1 & -1 & 0 & -1\\ 
1 & 1 & 0 & -1 & -1 & 0 & 0 & 0 & -1 & 0 & 0 & 1 & -1 & 0 & 0 & 0 & 1 & -1 & 0 & 0 & 0\\ 
-1 & 0 & 0 & -1 & -1 & -1 & 0 & 1 & 1 & 0 & 0 & 0 & 1 & -1 & 0 & 0 & 1 & 0 & 0 & 0 & 0\\ 
0 & -1 & -1 & 0 & 0 & 0 & 0 & 0 & 0 & 0 & 0 & 1 & 0 & -1 & 1 & 1 & -1 & -1 & 0 & 0 & 1

\end{array}$}
\]
\caption{Symmetric Weighing Matrix with n=21 w=9}
\label{fig:arrays2}
\end{figure*}

\begin{figure*}[h!]
\centering
\[
\fbox{
$\begin{array}{*{18}{r}}
0 & 0 & -1 & 0 & 0 & -1 & -1 & 0 & -1 & 0 & 0 & 1 & 1 & -1 & 1 & 1 & 0 & 0\\ 
0 & 0 & 1 & -1 & 0 & 1 & -1 & 1 & 0 & 0 & 0 & 0 & 1 & 1 & 0 & 1 & 0 & -1\\ 
1 & -1 & 0 & 0 & 0 & 0 & -1 & 1 & 1 & -1 & 0 & 0 & 0 & 0 & 1 & -1 & 0 & 1\\ 
0 & 1 & 0 & 0 & -1 & 1 & 1 & 0 & 0 & -1 & 0 & 1 & 1 & 0 & 0 & 0 & 1 & 1\\ 
0 & 0 & 0 & 1 & 0 & 0 & 0 & 1 & -1 & -1 & 1 & -1 & -1 & 0 & 0 & 1 & 1 & 0\\ 
1 & -1 & 0 & -1 & 0 & 0 & 1 & -1 & 0 & -1 & 1 & 0 & 0 & 0 & 0 & 1 & -1 & 0\\ 
1 & 1 & 1 & -1 & 0 & -1 & 0 & 0 & -1 & 0 & -1 & -1 & 0 & 0 & 0 & 0 & 0 & 1\\ 
0 & -1 & -1 & 0 & -1 & 1 & 0 & 0 & 0 & 1 & -1 & -1 & 0 & 0 & 0 & 1 & 0 & 1\\ 
1 & 0 & -1 & 0 & 1 & 0 & 1 & 0 & 0 & 0 & -1 & 0 & 0 & 1 & 1 & 0 & 1 & -1\\ 
0 & 0 & 1 & 1 & 1 & 1 & 0 & -1 & 0 & 0 & 0 & -1 & 1 & -1 & 1 & 0 & 0 & 0\\ 
0 & 0 & 0 & 0 & -1 & -1 & 1 & 1 & 1 & 0 & 0 & -1 & 1 & -1 & 0 & 0 & 0 & -1\\ 
-1 & 0 & 0 & -1 & 1 & 0 & 1 & 1 & 0 & 1 & 1 & 0 & 0 & 0 & 1 & 0 & 0 & 1\\ 
-1 & -1 & 0 & -1 & 1 & 0 & 0 & 0 & 0 & -1 & -1 & 0 & 0 & -1 & -1 & 0 & 1 & 0\\ 
1 & -1 & 0 & 0 & 0 & 0 & 0 & 0 & -1 & 1 & 1 & 0 & 1 & 0 & -1 & -1 & 1 & 0\\ 
-1 & 0 & -1 & 0 & 0 & 0 & 0 & 0 & -1 & -1 & 0 & -1 & 1 & 1 & 0 & -1 & -1 & 0\\ 
-1 & -1 & 1 & 0 & -1 & -1 & 0 & -1 & 0 & 0 & 0 & 0 & 0 & 1 & 1 & 0 & 1 & 0\\ 
0 & 0 & 0 & -1 & -1 & 1 & 0 & 0 & -1 & 0 & 0 & 0 & -1 & -1 & 1 & -1 & 0 & -1\\ 
0 & 1 & -1 & -1 & 0 & 0 & -1 & -1 & 1 & 0 & 1 & -1 & 0 & 0 & 0 & 0 & 1 & 0

\end{array}$}
\]
\caption{Skew Weighing Matrix with n=18 w=9}
\label{fig:arrays3}
\end{figure*}

\begin{figure*}[h!]
\centering
\[
\fbox{
$\begin{array}{*{17}{r}}
2 & 1 & 1 & 0 & 1 & 0 & 0 & 1 & 0 & 0 & 0 & 1 & 1 & 0 & 1 & 2 & 1\\ 
0 & 2 & 1 & 1 & 0 & 0 & 1 & 0 & 1 & 1 & 1 & 1 & 0 & 0 & 2 & 1 & 0\\ 
1 & 0 & 0 & 0 & 0 & 2 & 0 & 1 & 2 & 1 & 1 & 1 & 1 & 0 & 1 & 1 & 0\\ 
1 & 1 & 1 & 0 & 0 & 0 & 1 & 2 & 1 & 0 & 2 & 1 & 0 & 1 & 0 & 0 & 1\\ 
0 & 0 & 1 & 0 & 1 & 1 & 2 & 0 & 0 & 1 & 1 & 2 & 1 & 0 & 0 & 1 & 1\\ 
0 & 1 & 2 & 0 & 1 & 1 & 1 & 1 & 1 & 0 & 0 & 0 & 2 & 1 & 1 & 0 & 0\\ 
1 & 0 & 1 & 2 & 1 & 1 & 1 & 2 & 0 & 1 & 0 & 1 & 0 & 0 & 1 & 0 & 0\\ 
0 & 1 & 0 & 1 & 1 & 1 & 1 & 1 & 1 & 0 & 0 & 1 & 0 & 2 & 0 & 2 & 0\\ 
0 & 2 & 0 & 1 & 1 & 1 & 0 & 1 & 1 & 1 & 0 & 1 & 1 & 0 & 0 & 0 & 2\\ 
1 & 1 & 2 & 1 & 0 & 2 & 0 & 0 & 0 & 1 & 1 & 0 & 0 & 1 & 0 & 1 & 1\\ 
1 & 0 & 1 & 1 & 0 & 0 & 0 & 0 & 1 & 1 & 0 & 2 & 1 & 2 & 1 & 0 & 1\\ 
1 & 0 & 1 & 2 & 1 & 0 & 1 & 0 & 2 & 0 & 1 & 0 & 1 & 0 & 0 & 1 & 1\\ 
1 & 1 & 0 & 1 & 2 & 1 & 0 & 0 & 0 & 0 & 2 & 1 & 1 & 1 & 1 & 0 & 0\\ 
1 & 1 & 0 & 1 & 0 & 0 & 1 & 1 & 0 & 2 & 1 & 0 & 2 & 1 & 0 & 1 & 0\\ 
0 & 0 & 0 & 1 & 0 & 1 & 1 & 1 & 0 & 0 & 1 & 0 & 1 & 1 & 2 & 1 & 2\\ 
2 & 1 & 0 & 0 & 1 & 1 & 2 & 0 & 1 & 1 & 0 & 0 & 0 & 1 & 1 & 0 & 1\\ 
0 & 0 & 1 & 0 & 2 & 0 & 0 & 1 & 1 & 2 & 1 & 0 & 0 & 1 & 1 & 1 & 1

\end{array}$}
\]
\caption{Balanced Ternary Design with parameters (17,17;8,2,12;12,8)}
\label{fig:arrays12}
\end{figure*}

\begin{figure*}[h!]
\centering
\[
\fbox{
$\begin{array}{*{21}{r}}
1 & 2 & 1 & 0 & 1 & 2 & 0 & 2 & 1 & 0 & 0 & 0 & 0 & 0 & 0 & 1 & 0 & 1 & 0 & 0 & 0\\ 
0 & 0 & 0 & 1 & 1 & 0 & 2 & 2 & 1 & 2 & 0 & 1 & 0 & 1 & 0 & 0 & 0 & 0 & 0 & 1 & 0\\ 
0 & 0 & 2 & 0 & 2 & 1 & 0 & 0 & 0 & 1 & 0 & 1 & 0 & 0 & 2 & 0 & 0 & 0 & 1 & 1 & 1\\ 
2 & 0 & 0 & 0 & 0 & 1 & 1 & 0 & 1 & 0 & 0 & 0 & 1 & 2 & 2 & 0 & 0 & 1 & 0 & 1 & 0\\ 
2 & 0 & 1 & 0 & 0 & 0 & 0 & 1 & 0 & 1 & 0 & 2 & 2 & 0 & 0 & 1 & 1 & 0 & 1 & 0 & 0\\ 
0 & 0 & 1 & 0 & 1 & 1 & 2 & 0 & 0 & 0 & 2 & 0 & 1 & 1 & 0 & 2 & 0 & 0 & 1 & 0 & 0\\ 
1 & 0 & 0 & 2 & 0 & 2 & 1 & 0 & 0 & 1 & 0 & 0 & 0 & 0 & 0 & 0 & 1 & 1 & 2 & 0 & 1\\ 
0 & 2 & 0 & 0 & 0 & 0 & 0 & 0 & 1 & 1 & 0 & 0 & 1 & 1 & 0 & 1 & 0 & 0 & 2 & 2 & 1\\ 
1 & 0 & 0 & 0 & 1 & 0 & 0 & 1 & 0 & 1 & 2 & 0 & 1 & 0 & 0 & 0 & 0 & 2 & 0 & 1 & 2\\ 
0 & 0 & 0 & 2 & 1 & 0 & 0 & 0 & 2 & 0 & 0 & 1 & 1 & 0 & 1 & 2 & 0 & 1 & 0 & 0 & 1\\ 
0 & 0 & 0 & 0 & 0 & 1 & 0 & 1 & 2 & 0 & 2 & 1 & 0 & 0 & 1 & 0 & 2 & 0 & 1 & 1 & 0\\ 
0 & 0 & 2 & 1 & 0 & 0 & 0 & 1 & 0 & 0 & 0 & 0 & 0 & 2 & 0 & 1 & 2 & 1 & 0 & 1 & 1\\ 
0 & 2 & 0 & 0 & 1 & 0 & 2 & 0 & 0 & 0 & 0 & 1 & 1 & 0 & 1 & 0 & 2 & 1 & 0 & 0 & 1\\ 
1 & 2 & 1 & 2 & 0 & 0 & 0 & 0 & 0 & 1 & 2 & 1 & 0 & 1 & 1 & 0 & 0 & 0 & 0 & 0 & 0

\end{array}$}
\]
\caption{Balanced Ternary Design with parameters (14,21;6,3,12;8,6)}
\label{fig:arrays13}
\end{figure*}

\begin{figure*}[h!]
\centering
\[
\fbox{
$\begin{array}{*{16}{r}}
0 & 1 & 1 & 0 & 1 & 2 & 2 & 0 & 1 & 2 & 0 & 0 & 2 & 0 & 0 & 0\\ 
2 & 1 & 0 & 0 & 0 & 0 & 2 & 1 & 1 & 0 & 2 & 2 & 1 & 0 & 0 & 0\\ 
2 & 2 & 2 & 2 & 0 & 1 & 0 & 0 & 0 & 1 & 1 & 0 & 0 & 1 & 0 & 0\\ 
1 & 0 & 2 & 0 & 0 & 0 & 1 & 0 & 0 & 0 & 0 & 1 & 2 & 2 & 2 & 1\\ 
2 & 0 & 0 & 1 & 2 & 2 & 1 & 0 & 0 & 0 & 0 & 1 & 0 & 0 & 1 & 2\\ 
0 & 2 & 0 & 1 & 2 & 0 & 1 & 0 & 2 & 0 & 0 & 1 & 0 & 2 & 1 & 0\\ 
1 & 0 & 0 & 0 & 0 & 1 & 0 & 0 & 2 & 2 & 2 & 0 & 0 & 1 & 2 & 1\\ 
0 & 0 & 1 & 2 & 1 & 0 & 2 & 2 & 0 & 1 & 1 & 0 & 0 & 0 & 2 & 0\\ 
0 & 0 & 1 & 0 & 1 & 2 & 0 & 2 & 0 & 1 & 1 & 2 & 0 & 2 & 0 & 0\\ 
1 & 0 & 0 & 2 & 0 & 1 & 0 & 2 & 2 & 0 & 0 & 0 & 2 & 1 & 0 & 1\\ 
0 & 1 & 2 & 0 & 2 & 0 & 0 & 1 & 1 & 0 & 2 & 0 & 1 & 0 & 0 & 2\\ 
0 & 2 & 0 & 1 & 0 & 0 & 0 & 1 & 0 & 2 & 0 & 2 & 1 & 0 & 1 & 2

\end{array}$}
\]
\caption{Balanced Ternary Design with parameters (12,16;4,4,12;9,8)}
\label{fig:arrays14}
\end{figure*}

\begin{figure*}[h!]
\centering
\[
\fbox{
$\begin{array}{*{16}{r}}
1 & 1 & 1 & 0 & 2 & 1 & 0 & 0 & 0 & 1 & 1 & 0 & 2 & 2 & 1 & 0\\ 
0 & 2 & 1 & 1 & 0 & 0 & 0 & 1 & 1 & 2 & 2 & 0 & 0 & 1 & 1 & 1\\ 
2 & 2 & 1 & 1 & 1 & 0 & 1 & 2 & 0 & 0 & 1 & 1 & 1 & 0 & 0 & 0\\ 
1 & 0 & 1 & 0 & 2 & 0 & 0 & 1 & 1 & 1 & 1 & 2 & 0 & 1 & 0 & 2\\ 
0 & 0 & 1 & 2 & 1 & 0 & 1 & 2 & 1 & 0 & 0 & 0 & 1 & 2 & 1 & 1\\ 
2 & 1 & 0 & 2 & 1 & 1 & 0 & 0 & 1 & 1 & 0 & 0 & 1 & 0 & 1 & 2\\ 
0 & 1 & 0 & 1 & 0 & 1 & 1 & 1 & 0 & 2 & 0 & 2 & 2 & 1 & 0 & 1\\ 
2 & 1 & 0 & 0 & 0 & 1 & 1 & 1 & 2 & 1 & 0 & 1 & 0 & 2 & 1 & 0\\ 
0 & 1 & 0 & 1 & 2 & 0 & 2 & 0 & 2 & 1 & 1 & 1 & 1 & 0 & 1 & 0\\ 
1 & 0 & 2 & 1 & 1 & 1 & 1 & 1 & 0 & 2 & 0 & 1 & 0 & 0 & 2 & 0\\ 
0 & 1 & 0 & 0 & 1 & 2 & 0 & 2 & 1 & 0 & 1 & 1 & 1 & 0 & 2 & 1\\ 
1 & 0 & 0 & 1 & 1 & 2 & 2 & 1 & 0 & 1 & 2 & 0 & 0 & 1 & 0 & 1\\ 
0 & 2 & 2 & 1 & 1 & 2 & 1 & 0 & 1 & 0 & 0 & 1 & 0 & 1 & 0 & 1\\ 
1 & 0 & 1 & 2 & 0 & 1 & 0 & 0 & 1 & 0 & 2 & 2 & 1 & 1 & 1 & 0\\ 
1 & 0 & 2 & 0 & 0 & 1 & 1 & 1 & 2 & 1 & 1 & 0 & 2 & 0 & 0 & 1\\ 
1 & 1 & 1 & 0 & 0 & 0 & 2 & 0 & 0 & 0 & 1 & 1 & 1 & 1 & 2 & 2

\end{array}$}
\]
\caption{Balanced Ternary Design with parameters (16,16;7,3,13;13,10)}
\label{fig:arrays15}
\end{figure*}

\begin{figure*}[h!]
\centering
\[
\fbox{
$\begin{array}{*{21}{r}}
2 & 0 & 0 & 2 & 0 & 2 & 0 & 0 & 0 & 2 & 0 & 1 & 2 & 0 & 0 & 0 & 0 & 1 & 0 & 1 & 1\\ 
0 & 0 & 0 & 2 & 0 & 0 & 1 & 2 & 0 & 0 & 2 & 2 & 1 & 0 & 1 & 0 & 2 & 0 & 1 & 0 & 0\\ 
0 & 0 & 0 & 0 & 2 & 0 & 0 & 2 & 0 & 0 & 0 & 1 & 2 & 2 & 0 & 2 & 0 & 1 & 0 & 1 & 1\\ 
2 & 0 & 0 & 0 & 0 & 0 & 0 & 0 & 2 & 0 & 0 & 0 & 1 & 2 & 1 & 0 & 2 & 0 & 2 & 1 & 1\\ 
1 & 0 & 0 & 0 & 0 & 2 & 2 & 1 & 2 & 0 & 0 & 1 & 0 & 0 & 2 & 2 & 0 & 1 & 0 & 0 & 0\\ 
0 & 1 & 2 & 0 & 2 & 2 & 0 & 0 & 1 & 0 & 0 & 2 & 0 & 0 & 0 & 0 & 2 & 0 & 0 & 1 & 1\\ 
0 & 0 & 0 & 1 & 2 & 0 & 2 & 0 & 2 & 2 & 2 & 0 & 0 & 1 & 0 & 0 & 0 & 0 & 0 & 1 & 1\\ 
1 & 2 & 0 & 0 & 2 & 0 & 0 & 1 & 0 & 2 & 0 & 1 & 0 & 0 & 2 & 0 & 0 & 1 & 2 & 0 & 0\\ 
0 & 0 & 2 & 1 & 0 & 0 & 1 & 0 & 0 & 2 & 0 & 0 & 0 & 1 & 0 & 2 & 2 & 2 & 1 & 0 & 0\\ 
0 & 1 & 2 & 0 & 0 & 0 & 0 & 0 & 1 & 0 & 2 & 0 & 2 & 0 & 2 & 0 & 0 & 2 & 0 & 1 & 1\\ 
0 & 2 & 0 & 1 & 0 & 2 & 0 & 0 & 0 & 0 & 2 & 0 & 0 & 1 & 0 & 2 & 0 & 0 & 2 & 1 & 1\\ 
2 & 2 & 2 & 1 & 0 & 0 & 2 & 2 & 0 & 0 & 0 & 0 & 0 & 1 & 0 & 0 & 0 & 0 & 0 & 1 & 1

\end{array}$}
\]
\caption{Balanced Ternary Design with parameters (12,21;4,5,14;8,8)}
\label{fig:arrays16}
\end{figure*}

\begin{figure*}[h!]
\centering
\[
\fbox{
$\begin{array}{*{12}{r}}
6 & 5 & 7 & 9 & 1 & 4 & 11 & 2 & 10 & 8 & 0 & 3\\ 
0 & 9 & 2 & 8 & 1 & 4 & 11 & 3 & 10 & 6 & 7 & 5\\ 
4 & 5 & 7 & 8 & 1 & 0 & 6 & 3 & 10 & 2 & 9 & 11\\ 
6 & 7 & 3 & 8 & 1 & 0 & 11 & 9 & 10 & 5 & 2 & 4\\ 
6 & 5 & 0 & 8 & 9 & 7 & 11 & 3 & 10 & 1 & 4 & 2\\ 
6 & 5 & 2 & 8 & 1 & 9 & 7 & 11 & 10 & 4 & 3 & 0\\ 
4 & 5 & 3 & 0 & 1 & 9 & 11 & 7 & 10 & 6 & 8 & 2\\ 
0 & 5 & 4 & 6 & 1 & 7 & 11 & 9 & 10 & 2 & 3 & 8\\ 
7 & 5 & 1 & 8 & 4 & 0 & 11 & 2 & 10 & 6 & 3 & 9\\ 
4 & 6 & 9 & 8 & 1 & 7 & 11 & 2 & 10 & 3 & 5 & 0\\ 
0 & 5 & 7 & 8 & 6 & 9 & 11 & 4 & 10 & 3 & 2 & 1\\ 
3 & 5 & 7 & 8 & 1 & 10 & 11 & 9 & 2 & 6 & 4 & 0\\ 
7 & 5 & 8 & 4 & 1 & 6 & 11 & 3 & 10 & 9 & 2 & 0\\ 
4 & 5 & 2 & 8 & 0 & 6 & 11 & 9 & 10 & 7 & 1 & 3\\ 
9 & 5 & 3 & 8 & 7 & 4 & 11 & 1 & 10 & 2 & 6 & 0\\ 
7 & 5 & 6 & 8 & 1 & 4 & 0 & 9 & 10 & 3 & 11 & 2\\ 
0 & 5 & 3 & 8 & 1 & 6 & 9 & 2 & 10 & 11 & 4 & 7\\ 
9 & 5 & 2 & 7 & 1 & 0 & 11 & 8 & 10 & 3 & 4 & 6\\ 
9 & 5 & 11 & 8 & 1 & 7 & 4 & 0 & 10 & 6 & 2 & 3\\ 
7 & 0 & 5 & 8 & 1 & 9 & 11 & 6 & 10 & 2 & 4 & 3\\ 
9 & 4 & 7 & 8 & 1 & 6 & 11 & 5 & 10 & 0 & 3 & 2
\end{array}$}
\]
\caption{Equidistant Permutation Array with n=12 d=8 m=21}
\label{fig:arrays17}
\end{figure*}

\begin{figure*}[h!]
\centering
\[
\fbox{
$\begin{array}{*{20}{r}}
8 & 2 & 9 & 16 & 15 & 18 & 19 & 10 & 1 & 4 & 6 & 3 & 17 & 7 & 12 & 13 & 5 & 14 & 0 & 11\\ 
11 & 3 & 19 & 1 & 2 & 18 & 8 & 7 & 14 & 10 & 16 & 17 & 9 & 13 & 4 & 0 & 15 & 5 & 12 & 6\\ 
13 & 11 & 4 & 3 & 2 & 16 & 9 & 0 & 6 & 8 & 5 & 19 & 7 & 10 & 12 & 15 & 1 & 18 & 14 & 17\\ 
19 & 13 & 12 & 16 & 2 & 8 & 1 & 5 & 15 & 14 & 3 & 0 & 18 & 17 & 11 & 7 & 6 & 9 & 4 & 10\\ 
8 & 6 & 17 & 14 & 5 & 9 & 12 & 0 & 7 & 16 & 4 & 13 & 18 & 1 & 19 & 3 & 15 & 11 & 10 & 2\\ 
12 & 11 & 0 & 13 & 10 & 7 & 1 & 15 & 17 & 2 & 4 & 5 & 6 & 19 & 16 & 18 & 3 & 14 & 9 & 8\\ 
18 & 10 & 3 & 5 & 0 & 4 & 15 & 13 & 9 & 11 & 14 & 1 & 7 & 8 & 16 & 12 & 19 & 17 & 6 & 2
\end{array}$}
\]
\caption{Florentine Rectangle with r=7 n=20}
\label{fig:arrays18}
\end{figure*}

\begin{figure*}[h!]
\centering
\[
\fbox{
$\begin{array}{*{24}{r}}
6 & 13 & 17 & 3 & 20 & 10 & 5 & 16 & 8 & 15 & 21 & 22 & 0 & 18 & 4 & 14 & 7 & 12 & 11 & 23 & 19 & 1 & 9 & 2\\ 
14 & 2 & 20 & 6 & 0 & 4 & 21 & 23 & 11 & 17 & 8 & 3 & 18 & 7 & 19 & 10 & 13 & 22 & 15 & 16 & 5 & 1 & 12 & 9\\ 
14 & 5 & 3 & 19 & 20 & 1 & 11 & 7 & 10 & 0 & 6 & 23 & 21 & 2 & 16 & 9 & 22 & 13 & 12 & 8 & 4 & 15 & 18 & 17\\ 
16 & 15 & 2 & 1 & 3 & 21 & 0 & 17 & 13 & 10 & 19 & 9 & 18 & 12 & 4 & 11 & 6 & 5 & 8 & 7 & 22 & 23 & 20 & 14\\ 
2 & 9 & 7 & 23 & 18 & 15 & 20 & 13 & 14 & 1 & 5 & 4 & 3 & 8 & 22 & 12 & 6 & 21 & 11 & 19 & 0 & 16 & 10 & 17\\ 
19 & 23 & 12 & 18 & 5 & 22 & 16 & 4 & 9 & 17 & 10 & 7 & 3 & 1 & 2 & 0 & 20 & 15 & 8 & 6 & 11 & 14 & 13 & 21\\ 
8 & 0 & 13 & 11 & 9 & 15 & 22 & 19 & 5 & 17 & 23 & 2 & 6 & 14 & 12 & 1 & 10 & 18 & 21 & 7 & 4 & 20 & 3 & 16
\end{array}$}
\]
\caption{Florentine Rectangle with r=7 n=24}
\label{fig:arrays19}
\end{figure*}

\begin{figure*}[h!]
\centering
\[
\fbox{
$\begin{array}{*{25}{r}}
16 & 10 & 6 & 19 & 14 & 3 & 17 & 13 & 24 & 15 & 7 & 21 & 20 & 9 & 1 & 5 & 8 & 22 & 18 & 23 & 0 & 11 & 2 & 4 & 12\\ 
19 & 7 & 9 & 13 & 17 & 23 & 2 & 6 & 0 & 5 & 16 & 12 & 20 & 21 & 24 & 1 & 3 & 10 & 4 & 8 & 11 & 14 & 18 & 15 & 22\\ 
2 & 12 & 18 & 8 & 15 & 19 & 16 & 5 & 14 & 23 & 7 & 20 & 11 & 3 & 6 & 22 & 21 & 1 & 13 & 4 & 17 & 0 & 9 & 10 & 24\\ 
18 & 19 & 12 & 10 & 5 & 7 & 15 & 4 & 1 & 22 & 0 & 24 & 11 & 17 & 9 & 8 & 3 & 13 & 20 & 14 & 21 & 16 & 2 & 23 & 6\\ 
1 & 20 & 18 & 2 & 17 & 21 & 4 & 22 & 9 & 3 & 23 & 5 & 12 & 13 & 15 & 8 & 14 & 7 & 11 & 6 & 10 & 19 & 24 & 0 & 16\\ 
4 & 6 & 8 & 20 & 17 & 10 & 11 & 23 & 1 & 0 & 7 & 14 & 2 & 19 & 3 & 15 & 12 & 5 & 9 & 18 & 22 & 24 & 16 & 21 & 13\\ 
11 & 0 & 13 & 3 & 16 & 23 & 22 & 7 & 4 & 24 & 21 & 15 & 14 & 19 & 18 & 12 & 2 & 1 & 10 & 17 & 5 & 20 & 8 & 6 & 9
\end{array}$}
\]
\caption{Florentine Rectangle with r=7 n=25}
\label{fig:arrays20}
\end{figure*}

\begin{figure*}[h!]
\centering
\[
\fbox{
$\begin{array}{*{26}{r}}
6 & 14 & 25 & 12 & 4 & 17 & 13 & 9 & 15 & 11 & 8 & 23 & 7 & 19 & 10 & 21 & 3 & 2 & 18 & 22 & 16 & 24 & 20 & 5 & 0 & 1\\ 
17 & 24 & 14 & 10 & 1 & 2 & 19 & 15 & 25 & 11 & 13 & 16 & 4 & 6 & 3 & 5 & 9 & 12 & 23 & 20 & 0 & 18 & 8 & 7 & 22 & 21\\ 
17 & 16 & 25 & 4 & 14 & 21 & 10 & 18 & 6 & 12 & 1 & 5 & 24 & 22 & 19 & 13 & 7 & 0 & 20 & 2 & 15 & 3 & 23 & 8 & 11 & 9\\ 
19 & 22 & 12 & 17 & 11 & 7 & 24 & 9 & 1 & 18 & 25 & 21 & 5 & 3 & 8 & 14 & 0 & 15 & 23 & 10 & 6 & 4 & 16 & 2 & 13 & 20\\ 
21 & 0 & 14 & 15 & 6 & 19 & 9 & 3 & 20 & 7 & 25 & 2 & 22 & 5 & 1 & 13 & 10 & 8 & 12 & 24 & 16 & 11 & 17 & 4 & 18 & 23\\ 
9 & 5 & 10 & 3 & 18 & 1 & 4 & 20 & 25 & 16 & 23 & 24 & 21 & 15 & 22 & 13 & 11 & 2 & 14 & 17 & 7 & 12 & 0 & 19 & 6 & 8\\ 
7 & 20 & 22 & 8 & 16 & 5 & 2 & 6 & 24 & 17 & 23 & 15 & 18 & 3 & 19 & 0 & 10 & 4 & 9 & 11 & 12 & 21 & 13 & 25 & 14 & 1
\end{array}$}
\]
\caption{Florentine Rectangle with r=7 n=26}
\label{fig:arrays21}
\end{figure*}

\begin{figure*}[h!]
\centering
\[
\fbox{
$\begin{array}{*{27}{r}}
12 & 0 & 21 & 26 & 1 & 23 & 18 & 16 & 4 & 24 & 25 & 14 & 11 & 2 & 6 & 17 & 19 & 8 & 5 & 22 & 7 & 20 & 15 & 9 & 3 & 10 & 13\\ 
5 & 20 & 9 & 13 & 22 & 14 & 12 & 18 & 10 & 8 & 6 & 21 & 2 & 3 & 16 & 25 & 4 & 0 & 26 & 24 & 19 & 1 & 7 & 23 & 11 & 15 & 17\\ 
3 & 4 & 7 & 11 & 25 & 1 & 14 & 21 & 19 & 0 & 12 & 9 & 6 & 18 & 8 & 13 & 26 & 23 & 24 & 2 & 5 & 10 & 15 & 22 & 20 & 16 & 17\\ 
11 & 23 & 12 & 26 & 18 & 4 & 14 & 3 & 17 & 8 & 7 & 10 & 20 & 13 & 24 & 0 & 15 & 21 & 16 & 5 & 1 & 2 & 9 & 25 & 6 & 19 & 22\\ 
9 & 19 & 12 & 17 & 0 & 1 & 22 & 10 & 3 & 2 & 26 & 20 & 7 & 15 & 6 & 25 & 23 & 5 & 11 & 13 & 14 & 8 & 4 & 16 & 24 & 21 & 18\\ 
5 & 6 & 1 & 21 & 24 & 15 & 8 & 23 & 14 & 19 & 7 & 22 & 16 & 2 & 13 & 4 & 10 & 9 & 11 & 17 & 26 & 25 & 0 & 3 & 20 & 18 & 12\\ 
17 & 16 & 18 & 2 & 11 & 19 & 5 & 13 & 25 & 15 & 1 & 24 & 26 & 4 & 22 & 6 & 10 & 0 & 20 & 23 & 21 & 8 & 9 & 7 & 3 & 12 & 14
\end{array}$}
\]
\caption{Florentine Rectangle with r=7 n=27}
\label{fig:arrays22}
\end{figure*}

\end{document}